%% file: root.tex
\documentclass[letterpaper, 10 pt, conference]{ieeeconf}  %

\let\labelindent\relax

\IEEEoverridecommandlockouts                              %

\usepackage[pagebackref,breaklinks,colorlinks,citecolor=cvprblue]{hyperref}
\input{macros}

\title{\LARGE \bf
OoDIS: Anomaly Instance Segmentation and Detection Benchmark
}

\author{Alexey Nekrasov$^{1,\text{\faIcon[regular]{envelope}}}$, Rui Zhou$^{1,3}$, Miriam Ackermann$^2$, Alexander Hermans$^1$,\\ Bastian Leibe$^1$, Matthias Rottmann$^2$\\[3pt]
$^1$RWTH Aachen University (Germany) , $^2$IZMD, University of Wuppertal (Germany)\\
$^3$ Beijing Institute of Technology (China)
}

\begin{document}
\maketitle

\input{sections/0_abstract}

\let\thefootnote\relax\footnotetext{{\tt\small \faIcon[regular]{envelope}: nekrasov@vision.rwth-aachen.de}}

\input{sections/1_introduction}

\input{sections/2_related_work}

\input{sections/3_method}

\input{sections/4_results}

\input{sections/5_conclusion}

\section*{Acknowledgment}
We acknowledge fruitful discussions with H.\ Blum, R.\ Chan, S.\ Gasperini and S.\ Rai, as well as a contribution of annotations for Fishyscapes by H.\ Blum, and help with the benchmark submission from S.\ Gasperini and S.\ Rai.
M.R.\ \& M.A.\ acknowledge support by the German Federal Ministry of Education and Research (BMBF) within the junior research group project ``UnrEAL'' (grant no.\ 01IS22069).
A.N.\ acknowledges funding by the BMBF project ``WestAI'' (grant no.\ 01IS22094D).
We thank M.\ Burdorf, G.\ Lydakis, C.\ Schmidt, and others in the lab for discussions and feedback.

{
    \bibliographystyle{IEEEtran}
    \bibliography{main}
}

\thispagestyle{empty}
\pagestyle{empty}
\end{document}

%% file: macros.tex
\usepackage{bm}

\usepackage{xspace}

\usepackage{array}
\usepackage{booktabs}
\usepackage{fontawesome5}
\usepackage{xcolor,colortbl}
\definecolor{Gray}{gray}{0.9}
\definecolor{Textgray}{gray}{0.4}

\definecolor{notetext}{rgb}{0.7,0,0}
\definecolor{notetext}{rgb}{0.7,0,0}

\usepackage[nointegrals]{wasysym}
\usepackage{makecell}
\usepackage{balance}
\usepackage{multirow}
\usepackage{overpic}
\usepackage{bbm}
\usepackage{dsfont}
\usepackage[font=small,labelfont=bf]{caption}
\usepackage{color}
\usepackage{pifont}
\usepackage{etoolbox}
\usepackage{float}
\usepackage{tikz}
\usepackage{svg}
\usepackage{times}
\usepackage{epsfig}
\usepackage{graphicx}
\usepackage{amsmath}
\usepackage{amssymb}
\usepackage{comment}
\usepackage{soul}
\usepackage{listings}
\usepackage{lscape}
\usepackage{rotating}
\usepackage{enumitem}
\usepackage[utf8]{inputenc}
\usepackage{pgfplots}
\usepackage{import}
\usepackage{xifthen}
\usepackage{transparent}
\usepackage{tabularx}
\usepackage{tabu}
\usepackage{arydshln}
\DeclareUnicodeCharacter{2212}{−}
\usepgfplotslibrary{groupplots,dateplot,statistics}
\usetikzlibrary{patterns,shapes.arrows}
\pgfplotsset{compat=newest}
\usepackage[symbol]{footmisc}

\usepackage{subcaption}
\usepackage{xfrac}
\usepackage{tcolorbox}

\definecolor{LightCyan}{rgb}{0.87,0.92,0.96}
\definecolor{m_darkgreen}{HTML}{d5e9d5}
\definecolor{m_orange}{HTML}{fff2cd}
\definecolor{m_red}{HTML}{fccecd}
\definecolor{m_violet}{HTML}{e2d6e8}
\definecolor{m_blue}{HTML}{d8e9fc}

\definecolor{m_red}{RGB}{255,209,209}
\definecolor{m_red_border}{RGB}{215,23,20}
\definecolor{m_orange}{RGB}{226,213,231}
\definecolor{m_orange_border}{RGB}{150,114,164}
\definecolor{m_blue}{RGB}{217,232,251}
\definecolor{m_blue_border}{RGB}{107,141,190}
\definecolor{m_yellow}{RGB}{255,242,205}
\definecolor{m_yellow_border}{RGB}{213,182,82}
\definecolor{m_gray}{RGB}{245,245,245}
\definecolor{m_gray_border}{RGB}{102,102,102}
\definecolor{m_green}{HTML}{d5e9d5}
\definecolor{m_green_border}{HTML}{84B158}

\lstdefinestyle{mypython}{
  language=python,
  breaklines=true,
  basicstyle=\fontsize{8}{12}\selectfont\ttfamily,
  keywordstyle=\bfseries\color{my_blue},
  linewidth=.99\textwidth,
}

\DeclareRobustCommand{\colorsquare}[1]{\tikz{\path[draw=#1_border,fill=#1, thick, rounded corners=0.6pt] (0,0) rectangle (6pt,6pt);}}

\newcommand{\parag}[1]{\vskip4pt \noindent \textbf{#1}}

\newcommand{\PARbegin}[1]{\noindent {\bf #1~}}

\usepackage[capitalize]{cleveref}
\crefname{section}{Sec.}{Secs.}
\Crefname{section}{Section}{Sections}
\Crefname{table}{Table}{Tables}
\crefname{table}{Tab.}{Tabs.}

\definecolor{darkgreen}{RGB}{0,255,0}
\definecolor{linkgreen}{RGB}{52,130,48}

\newcommand{\cmark}{\ding{51}}%
\newcommand{\xmark}{\ding{55}}%

\newcolumntype{Y}{>{\centering\arraybackslash}X}
\newcolumntype{Z}{>{\raggedleft\arraybackslash}X}

\newif\ifmynotes
\mynotestrue
\definecolor{notetext}{rgb}{0.7,0,0}

\makeatletter
\def\adl@drawiv#1#2#3{%
        \hskip.5\tabcolsep
        \xleaders#3{#2.5\@tempdimb #1{1}#2.5\@tempdimb}%
                #2\z@ plus1fil minus1fil\relax
        \hskip.5\tabcolsep}
\newcommand{\cdashlinelr}[1]{%
  \noalign{\vskip\aboverulesep
           \global\let\@dashdrawstore\adl@draw
           \global\let\adl@draw\adl@drawiv}
  \cdashline{#1}
  \noalign{\global\let\adl@draw\@dashdrawstore
           \vskip\belowrulesep}}
\makeatother

%% file: sections/0_abstract.tex
\begin{abstract}

    Safe navigation of self-driving cars and robots requires a precise understanding of their environment.
    Training data for perception systems cannot cover the wide variety of objects that may appear during deployment.
    Thus, reliable identification of unknown objects, such as wild animals and untypical obstacles, is critical due to their potential to cause serious accidents.
    Significant progress in semantic segmentation of anomalies has been facilitated by the availability of out-of-distribution (OOD) benchmarks.
    However, a comprehensive understanding of scene dynamics requires the segmentation of individual objects, and thus the segmentation of instances is essential.
    Development in this area has been lagging, largely due to the lack of dedicated benchmarks.
    The situation is similar in object detection.
    While there is interest in detecting and potentially tracking every anomalous object, the availability of dedicated benchmarks is clearly limited.
    To address this gap, this work extends some commonly used anomaly segmentation benchmarks to include the instance segmentation and object detection tasks.
    Our evaluation of anomaly instance segmentation and object detection methods shows that both of these challenges remain unsolved problems.
    We provide a competition and benchmark website under \url{https://vision.rwth-aachen.de/oodis}.
\end{abstract}

%% file: sections/1_introduction.tex
\section{Introduction}
\label{sec:intro}

Modern segmentation methods~\cite{cheng2022mask2former,chen2018deeplabv3} as well as object detection methods \cite{yolov8} perform well on curated closed-world datasets with a fixed set of classes.
However, models trained with a fixed training set fall short of solving the task when unexpected objects are present~\cite{hein2019relu,hendrycks2018baseline}.
These anomalies often cause models to misclassify, assigning known classes to unknown objects~\cite{guo2017calibration,kendall2017what}.
To prevent such behavior in real world applications, it is important to design or adapt models to handle such anomalies.
The task of anomaly detection spans multiple modalities~\cite{wong2019osis,bergmann2022mvtec3d,park2020video,maag2022two}, applications~\cite{bergmann2019mvtec,li2022coda}, and tasks~\cite{du2021vos,yang2022openood,tian2022pebal}.
The particular focus of this work is two-fold.
We focus on 1) the anomaly instance segmentation task, that aims to provide segmentation models with the ability to segment out-of-distribution (OOD) objects, and closely related to that 2) the anomaly object detection task, that aims to provide bounding boxes for OOD objects.
Both tasks are particularly critical for autonomous driving scenarios, where a recognition error can cause serious accidents.
A collision with lost cargo on the road or with livestock could be life-threatening.
To evaluate the performance of anomaly segmentation methods, a number of benchmarks have been proposed \cite{pinggera2016laf,chan2021segmentmeifyoucan}.

\begin{figure}
    \centering
    \includegraphics[width=\linewidth]{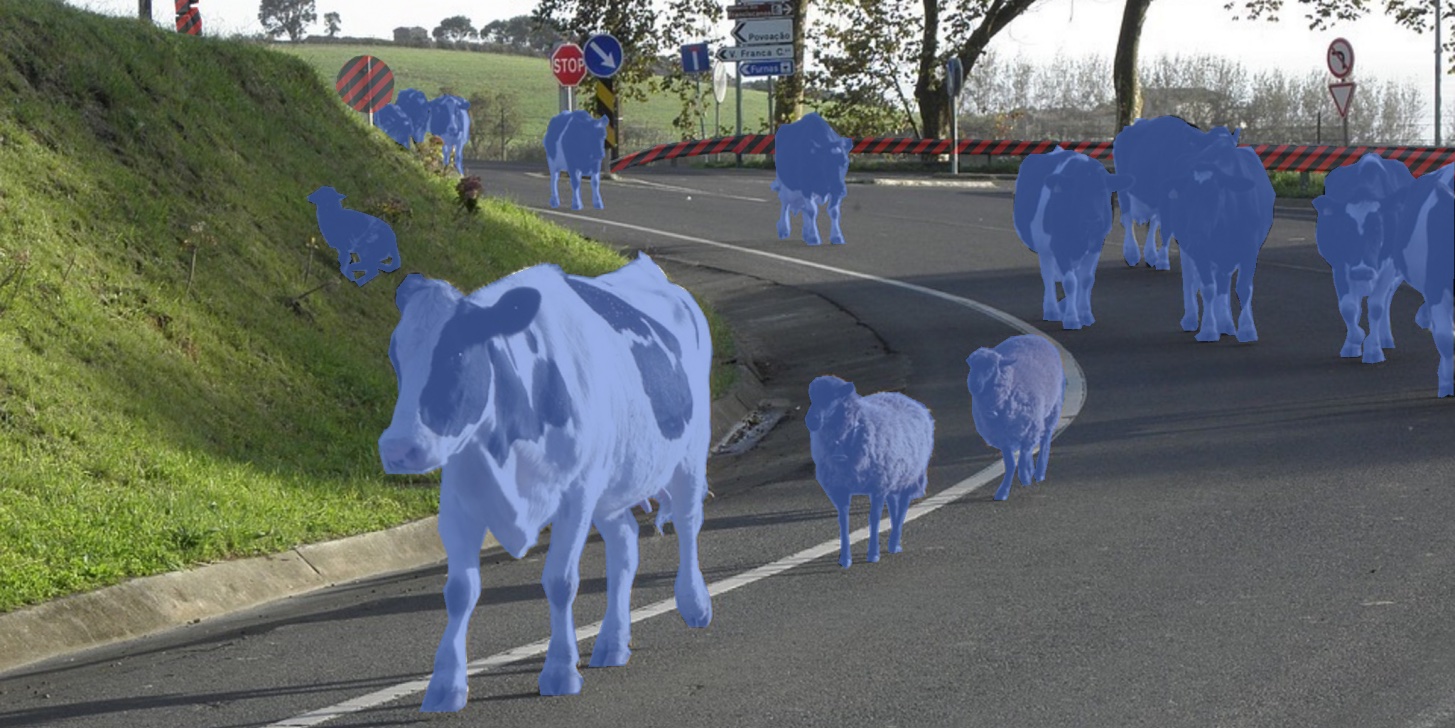}
    \includegraphics[width=\linewidth]{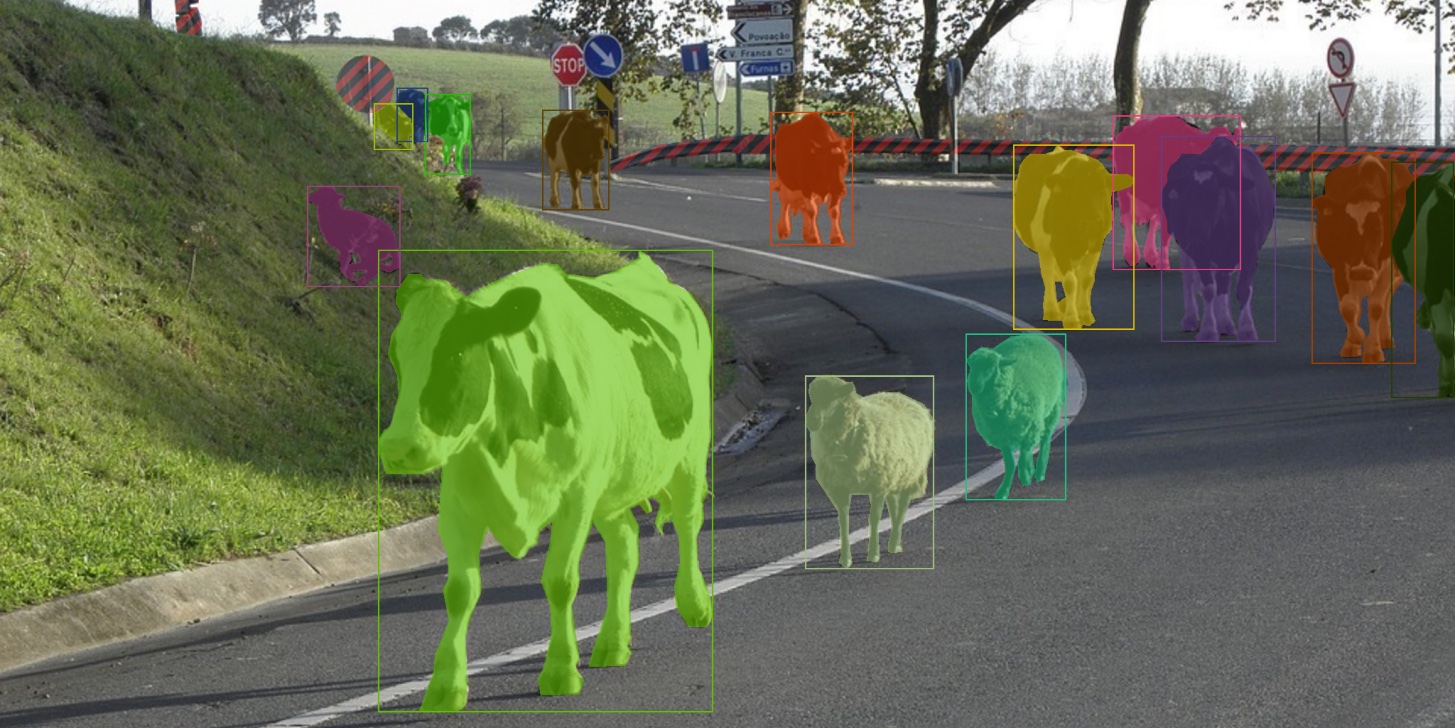} 
    \caption{
    Annotation example for the previous semantic annotation of the RoadAnomaly21 dataset (top) and the extended annotation labels (bottom) for our newly proposed benchmark.
    Accurate instance level bounding boxes and segmentation masks are provided as part of OoDIS to evaluate methods for anomaly instance segmentation as well as anomalous object detection.
    }
    \label{fig:example1}
\end{figure}

While anomaly segmentation~\cite{nayal2023rba,tian2022pebal,liang2022gmmseg} methods achieve exciting results on popular benchmarks, the area of anomaly instance segmentation remains unexplored.
Early datasets~\cite{pinggera2016laf} for anomaly segmentation included partial instance annotations of anomalies, but recently proposed datasets omit instance information~\cite{blum2021fishyscapes,chan2021segmentmeifyoucan}.
However, instance segmentation is critical for understanding complex scenes with multiple anomalous objects, such as cows and sheep as shown in Figure \ref{fig:example1}, that may appear in a group.
Previous anomaly segmentation approaches that operate on a pixel level would fail to distinguish individual objects.
Understanding these objects separately provides context about the potential dynamics of a scene, improving downstream tasks such as navigation or planning.
We hypothesize that recent advances in open set~\cite{wong2019osis,hwang2021eopsn} and class-agnostic~\cite{kirillov2023sam} instance segmentation have encouraged research in the area of anomaly instance segmentation, which was previously too challenging.
Recently, three works following different paradigms proposed to solve the task of anomaly instance segmentation~\cite{gasperini2023u3hs,rai2023mask2anomaly,nekrasov2023ugains}.
However, each of these works proposes a different evaluation procedure.
The situation is similar in object detection.
An in-distribution (ID) and OOD combination of datasets used for the evaluation of anomalous object detection is PascalVOC and \cite{everingham2015pascal} and COCO \cite{lin2014microsoft}.
The classes available in COCO but not in PascalVOC serve as OOD classes, see \cite{du2021vos}.
In anomaly object detection, a challenging and unified evaluation procedure will be beneficial for the development of the field.

To address the outlined limitations, we propose two benchmarks and evaluate existing anomalous instance segmentation and object detection methods in a unified manner, respectively.
We extend the labels of popular anomaly segmentation datasets~\cite{blum2021fishyscapes,chan2021segmentmeifyoucan} to instance segmentation, which naturally includes bounding boxes.
These datasets provide diverse real-world cases of road anomalies with precise annotations.
We reuse the Average Precision (AP) metric~\cite{hariharan2014ap} for instance evaluation similarly to the Cityscapes setup~\cite{cordts2016cityscapes}, with a slight modification to evaluate instances as small as 10 pixels in size.
In comparison to the semantic anomaly benchmarks, the AP metric in both tasks avoids size bias and requires high precision for smaller anomalous objects.
This is particularly important in the context of autonomous driving, where detecting anomalies in the distance is critical to give the system time to react.

To this end, we re-annotated anomalies within the Fishyscapes~\cite{blum2021fishyscapes}, RoadAnomaly21, and RoadObstacle21~\cite{chan2021segmentmeifyoucan} datasets
to evaluate anomaly instance segmentation and anomalous object detection methods.
We apply publicly available instance segmentation and object detection methods on both validation and test set and provide quantitative evaluation of the results.
Our evaluations show that while current anomaly segmentation methods perform well on semantic anomaly segmentation, instance segmentation methods only achieve moderate performance, suggesting a considerable space for improvement. In anomalous object detection, currently available methods exhibit a clear deficiency in detecting anomalous objects in challenging environmental conditions. This holds also true for more recent open-vocabulary object detectors such as \cite{liu2023grounding}.
We make validation data available on our challenge website, and open a submission portal where new approaches can be submitted.

%% file: sections/2_related_work.tex
\section{Related Work}
\label{sec:relwork}
\PARbegin{Out-of-Distribution (OOD) Datasets} have primarily focused on classification tasks, with several benchmarks recently introduced~\cite{yang2022openood,zhang2023openood}.
A common evaluation task is disentanglement of two classification datasets such as CIFAR and SVHN.
Methods such as deep ensembles~\cite{lakshminarayanan2017deepensemble} and Monte Carlo dropout~\cite{srivastava2014mcdropout}, while performing well on OOD classification, show limited usefulness in anomaly segmentation tasks~\cite{chan2021segmentmeifyoucan}.
Open-set instance segmentation~\cite{hwang2021eopsn,wong2019osis} assumes the presence of OOD data during training, a condition not applicable to anomaly segmentation where completely unseen objects may appear~\cite{gasperini2023u3hs}.
In autonomous driving, novel evaluation schemes have been proposed for detection tasks~\cite{du2021vos,li2022coda}.
However, these works do not address the need for precise pixel-level mapping in monocular driving detection setups.
Our work explores the segmentation of anomaly instances, which allows accurate prediction of individual, previously unseen, objects.

\parag{Anomaly Segmentation Datasets.}
Anomaly segmentation has received significant attention with the emergence of several recent datasets and benchmarks~\cite{pinggera2016laf,blum2021fishyscapes,chan2021segmentmeifyoucan}.
The Lost and Found (L\&F) dataset~\cite{pinggera2016laf} introduced the task of anomaly segmentation in a camera setup similar to the one used for the Cityscapes dataset~\cite{cordts2016cityscapes}.
L\&F has annotations limited to the road area and anomaly classes; however, it has questionable labels that include bicycles and kids as anomalies~\cite{blum2021fishyscapes}.
To fully control for anomalies in the training and test sets, the CAOS benchmark~\cite{hendrycks2022scaling} introduces a real dataset based on BDD100K~\cite{yu2020bdd100k}, treating certain inlier classes as anomalies, and a synthetic dataset for training and testing.
FishyScapes Lost and Found (FS L\&F)~\cite{blum2021fishyscapes} reannotates images from L\&F to extend in-distribution regions outside of the road class and introduces a separate benchmark with artificial anomalies.
Despite its popularity, FS L\&F lacks anomaly instance segmentation and it is constrained to lost cargo on the road.
To solve the diversity issue, SegmentMeIfYouCan~\cite{chan2021segmentmeifyoucan} introduces a diverse dataset with real anomalies on roads, which are not limited to the Cityscapes camera perspective.
In past years, evaluation on FS L\&F and SegmentMeIfYouCan dataset has been a standard practice.
However, instance annotations are missing from these datasets. 
Our work aims to extend these popular benchmarks by providing accurate instance annotations.

\parag{Anomaly Segmentation Methods.}
Segmentation of anomaly instances has been underexplored until recently.
There are previous works in open-set instance segmentation~\cite{wong2019osis,hwang2021eopsn}. However, they rely on unknown objects present in the training set; and methods that rely on depth cues~\cite{singh2020lidar} that are not applicable in general case.
In general anomaly instance segmentation methods produce per-pixel anomaly scores, while providing anomaly instances too.
U3HS~\cite{gasperini2023u3hs} uses uncertainty in semantic predictions to guide the region segmentation, and then clusters predicted class-agnostic instance embeddings.
Mask2Anomaly~\cite{rai2023mask2anomaly} applies modifications to the Mask2Former~\cite{cheng2022mask2former} architecture to produce reliable semantic anomaly scores in background regions, and uses a connected components on anomaly scores with a strategy to remove false-positives using intersections with in-distribution predictions.
UGainS~\cite{nekrasov2023ugains} combines the RbA anomaly segmentation method~\cite{nayal2023rba} with an interactive segmentation model~\cite{kirillov2023sam} to predict instances using point prompting.
Given the limited number of specialized methods for anomaly instance segmentation, we evaluate these models and analyze their performance, offering insights into their practical applications and limitations.

\parag{Anomalous Object Detection Datasets \& Methods.}
Up to now, compared to the amount of existing anomaly segmentation benchmarks, the availability of dedicated ID-OOD dataset combinations for object detection is limited.
The CODA benchmark \cite{li2022coda} introduces corner case datasets that contain rare scenes that naturally occur during driving. 
With about 90\% of the recorded corner cases claimed as novel.
Compared to the CODA benchmark, our benchmark proposed in the present work contains rather remote OOD objects, atypical for street scenes, while CODA contains rather near OOD objects that can be found comparatively frequently.

In \cite{du2021vos} (VOS), PascalVOC \cite{everingham2015pascal} is used as ID and complemented with COCO \cite{lin2014microsoft}, where the classes from COCO that are not in PascalVOC serve as OOD classes.
Everyday scenes are not safety-critical per-se.
In \cite{wilson2023safe} (SAFE), BDD100k serve as ID and COCO as well as PascalVOC as OOD, however the task in \cite{wilson2023safe} is to find OOD data rather than OOD objects.
In light of these experimental setups, the lack of dedicated ID-OOD dataset combinations in safety-critical contexts becomes apparent, which is addressed by the present work.
In a recent work \cite{ilyas2024potentialopenvocabularymodelsobject} based on a preprint of the present work \cite{nekrasov2024oodisanomalyinstancesegmentation} it turned out that open-vocabulary object detectors such as grounding Dino \cite{liu2023grounding} show clearly superior performance over conventional anomalous object detection methods such as VOS. However, it also turns out that there is still plenty of room for further improvement. A summary of the related dataset landscape is provided in \cref{tab:datasets}.

\input{tables/comparison}

%% file: tables/comparison.tex
\begin{table*}[t]
\caption{Comparison to previous anomaly benchmarks. The number of connected components are stated with respect to the test set. (*): For the CODA dataset, about 90\% of the instances are assessed to be anomalous in \cite{li2022coda}.}
\label{tab:datasets}
\resizebox{\textwidth}{!}{%
\begin{tabular}{@{}lccccccccc@{}}
\toprule
\textbf{Dataset} & \textbf{Year} & \textbf{Size} & \textbf{Anomaly} & \textbf{No.\ of} & \textbf{Sequences} & \textbf{Instance} & \textbf{Bounding} & \textbf{Private} & \textbf{Benchmark}  \\ 
& & \textbf{(Test/Val)} & \textbf{Source} & \textbf{Components} &  \textbf{(\#)} & \textbf{Labels} & \textbf{Boxes} & & \\
\midrule
\multicolumn{9}{l}{\textbf{Wuppertal OoD Tracking}}\\
Street Obstacle Sequences & 2022 & 1129 / - & Staged Obj. & 1592 & \cmark (20) & \cmark & \xmark & \xmark & instance tracking \\
Carla Wildlife & 2022 & 1210 / - & Simulated / Staged Obj. & 2826 & \cmark (26) & \cmark & \xmark & \xmark & instance tracking \\
Wuppertal Obstacle Sequences & 2022 & 938 / - & Staged Obj. & 1039 & \cmark (44) & \cmark & \xmark & \xmark & instance tracking \\
\midrule
\multicolumn{9}{l}{\textbf{CODA Corner Case Dataset}}\\
CODA-KITTI & 2022 & 309 / - & Natural Occurance & 399${}^{(*)}$ & subsampled & \xmark & \cmark & \xmark & object detection \\
CODA-nuScenes & 2022 & 134 / - & Natural Occurance & 1125${}^{(*)}$ & subsampled & \xmark & \cmark & \xmark & object detection \\
CODA-ONCE & 2022 & 1057 / - & Natural Occurance & 4413${}^{(*)}$ & subsampled  & \xmark & \cmark & \xmark & object detection \\
\midrule
\multicolumn{9}{l}{\textbf{Fishyscapes}} \\
FS Lost \& Found & 2019 & 275 / 100 & Staged Obj. & 165(?) & subsampled & \xmark & \xmark & \cmark & semantic segmentation \\
\midrule
\multicolumn{9}{l}{\textbf{SegmentMeIfYouCan}} \\
RoadAnomaly21 & 2021 & 100 / 10 & Web Sourcing & 262 & \xmark & \xmark & \xmark & \cmark & semantic segmentation\\
RoadObstacle21 & 2021 & 412 / 30 & Staged Obj. & 388 & subsampled & \xmark & \xmark & \cmark & semantic segmentation \\ 
\midrule
\multicolumn{9}{l}{\textbf{OoDIS}} \\
OoDIS & 2024 & 787 / 140 & Combined & 1735 & subsampled   & \cmark & \cmark & \cmark & instance segmentation \\ 
 & & & & & \& \xmark &  & & &  \& object detection \\
\bottomrule
\end{tabular}%
}
\end{table*}

%% file: sections/3_method.tex
\section{Benchmark and Evaluated Methods}
Anomaly segmentation as a task attempts to identify unexpected objects unknown during training.
Common examples include a deer or a cardboard box that may appear in the middle of the road.
Per-pixel segmentation does not provide sufficient information for downstream tasks such as tracking or navigation.
The challenging problems of instance segmentation and object detection remain under-explored and lack accessible benchmarks.
This benchmark addresses the lack of test evaluation protocols available to the community.

We aim to fill the gap by extending the labels of SegmentMeIfYouCan~\cite{chan2021segmentmeifyoucan} and FS L\&F~\cite{blum2021fishyscapes} datasets for instance segmentation, which naturally includes object detection.
We merge these datasets into a unified benchmark and adopt commonly used Average Precision (AP) metrics~\cite{lin2014microsoft}, that closely follows the Cityscapes~\cite{cordts2016cityscapes} segmentation benchmark.

\parag{Data.}
We use three datasets for anomaly segmentation and detection: RoadAnomaly21 and RoadObstacle21 from SegmentMeIfYouCan~\cite{chan2021segmentmeifyoucan}, and FS L\&F~\cite{blum2021fishyscapes}.
These are the standard benchmarks for the task, and they complement each other in label diversity well (see Figure \ref{fig:mask-distribution}).
To maintain data integrity, we keep the test sets from the datasets intact, using $100$ images from RoadAnomaly21, $412$ from RoadObstacle21, and $275$ from FS L\&F as our full test sets.
In addition, we provide a relabeled validation set of 100 images from FS L\&F.

The test set contains three relabeled datasets with different properties, but shares a common in-distribution dataset.
For the submission to the benchmark, we allow models trained on $19$ Cityscapes~\cite{cordts2016cityscapes} classes as the in-distribution dataset, and allow the use of auxiliary data, such as COCO~\cite{lin2014microsoft} to introduce virtual anomalies, similar to other anomaly segmentation works~\cite{grcic2022densehybrid,nayal2023rba,tian2022pebal,dibiase2021synboost,chan2021entropy,grcic2023m2feam}.
It is important to note that we expect no explicit supervision to segment unknowns, much like in the real world, it is a priori unknown know what kind of anomalies a given vision system will encounter.

The benchmark data contains three classes: inlier, outlier, and ignore. 
In-distribution regions contain classes known to Cityscapes; ignore regions are ambiguous regions that neither contain anomalies nor are in-distribution regions; and the outlier class contains anomalous instances (see Figure \ref{fig:example1}).
Ignore regions are ambiguous regions for which a class cannot be defined; common cases in Cityscapes are: bridges, advertisement posts, back sides of street signs and dark regions where the class could not be determined.
We omit ignore regions in evaluation and discard cases that overlap significantly with these regions.
We evaluate predictions only for the outlier class, without focusing on evaluation of in-distribution predictions.
To calculate the final Average Precision (AP) score, we compute a weighted average based on the number of images in each dataset.

Altogether, our instance segmentation versions of Fishyscapes L\&F, RoadAnomaly21 and RoadObstacle21 provide a diverse setting for numerical experiments.
The objects vary considerably in their size and, in particular in RoadAnomaly, the scenes have strongly varying numbers of instances, see \cref{fig:mask-distribution}.

\input{figures/mask_area}

\parag{Labeling Policy.}
In RoadAnomaly21, anomalies are of arbitrary size, located anywhere on the image, containing highly diverse samples.
Each individual object, such as an animal or object, is labeled as an individual object without introducing group labels.
FS L\&F mainly contains anomalies on the road, separate objects such as stacked boxes, which are treated as separate instances.
Only ambiguous regions are treated as ignore for RoadAnomaly21 and FS L\&F.
For RoadObstacle21, however, only the drivable area is considered an inlier, and everything outside the drivable area, including anomalies, are labeled as ignore regions.
Gaps within complex anomalies are also treated as ignore regions.
Each labeled object on an image is given a unique identifier.
Bounding boxes are also generated to facilitate anomaly localization and allow for the evaluation of anomaly object detection.

\parag{Metrics.}
Conventional anomaly segmentation metrics tend to favor larger objects.
Per pixel Average Precision (AP) or False Positive Rate (FPR) metrics, or sIoU \cite{chan2021segmentmeifyoucan}, which groups anomalies together, do not provide the correct evaluation metric.
We utilize the Cityscapes instance segmentation \cite{cordts2016cityscapes} evaluation suite for anomaly instance segmentation as well as the COCO evaluation suite \cite{lin2014microsoft} for anomalous object detection. Our benchmark primarily uses the AP metric (not the be confused with the per-pixel AP), a standard in instance segmentation and object detection. We describe it briefly in the following:

We assume that a prediction comes with a confidence score $s$. Given the ground truth, a prediction is assessed for correctness by applying a threshold to the IoU \cite{Jaccard12similarityCoefficient}. Up to technical details, for a given IoU threshold $t\%$, the precision recall curve is computed by varying the confidence $s$. The area under that curve is referred to as AP$t$. We consider AP50 ($t=50$) and the AP, which is an average over different thresholds, namely $\mathrm{AP}= \frac{1}{T} \sum_{t\in T} \mathrm{AP}t $ for  $T=\{50,55,60,\ldots,95\}$. This set of thresholds is used in both evaluation protocols, Cityscapes and COCO.

As additional evaluation metric, we provide the average recall (AR). Restricting the number of predictions to $k$, only keeping predictions with highest confidence $s$, the corresponding recall $\mathrm{REC}t(k)$ is averaged over a chosen set $T$ of IoU thresholds, i.e., $\mathrm{AR}k = \frac{1}{|T|}\sum_{t \in T} \mathrm{REC}t(k)$.
We provide AR1, AR10 and AR100.
The recall reflects how many of the anomalous objects can be segmented / detected and how many of them are overlooked.
Finally, we also report the number of the Predictions Per Frame (PPF) averaged over all images of the respective dataset, inspired by \cite{jena2023map}. This metric detects an over-production of predictions, e.g.\ aiming at matching the ground truth by chance.

\parag{Methods for Anomaly Instance Segmentation and Anomalous Object Detection.}

The \textbf{U3HS}~\cite{gasperini2023u3hs} method belongs to a class of models that neither require auxiliary data nor external models for instance segmentation.
The core of the method is the ability to learn class-agnostic instance embeddings that generalize beyond the training distribution.
These embeddings in uncertain regions are clustered to get instance predictions.
This allows clustering of anomalous regions occluded by other objects.

\textbf{Mask2Anomaly}~\cite{rai2023mask2anomaly} is a model that uses auxilary data, but does not use an external model for instance segmentation.
Common to other methods in the community~\cite{grcic2022densehybrid,tian2022pebal}, the model uses auxiliary data from COCO~\cite{lin2014microsoft} for guiding the anomaly scores that are grouped using connected components to form instance proposals.
To reduce the number of false positives, Mask2Anomaly introduces a post-processing strategy.
It computes the intersection with predicted in-distribution masks and uses class entropy to determine true instance proposals.

\textbf{UGainS}~\cite{nekrasov2023ugains} is a method that uses both auxiliary data and an external generalist segmentation model, namely the segment anything model (SAM)~\cite{kirillov2023sam}.
The method uses the anomaly segmentation method RbA~\cite{nayal2023rba} based on Mask2Former~\cite{cheng2022mask2former}, fine-tuned using data from COCO, to generate uncertainty regions.
UGainS uses farthest point sampling to sample a number of points from these regions as prompts for SAM~\cite{kirillov2023sam}.

%% file: figures/mask_area.tex
\begin{figure}
    \centering
    \resizebox{0.22\textwidth}{!}{\input{figures/mask_area.pgf}}
    \resizebox{0.22\textwidth}{!}{\input{figures/num_instances.pgf}}
    \caption{
    \textbf{Diversity of instance labels.}
    RodAnomaly21~\colorsquare{m_blue} typically contains multiple objects, while RoadObstacle21~\colorsquare{m_yellow} contains smaller objects in smaller quantities, and Fishyscapes L\&F~\colorsquare{m_green} provides a balance between the two.
    }
    \label{fig:mask-distribution}
\end{figure}

%% file: figures/mask_area.pgf
\begin{tikzpicture}

\definecolor{darkgray176}{RGB}{176,176,176}
\definecolor{sandybrown25520258}{RGB}{255,202,58}
\definecolor{steelblue25130196}{RGB}{25,130,196}
\definecolor{yellowgreen13820138}{RGB}{138,201,38}

\begin{axis}[
log basis x={10},
tick align=outside,
tick pos=left,
x grid style={darkgray176},
xlabel={Mask Size},
xmin=1.00237446725454, xmax=3990524.62993776,
xmode=log,
xtick style={color=black},
xtick={0.1,1,10,100,1000,10000,100000,1000000,10000000,100000000},
xticklabels={
  \(\displaystyle {10^{-1}}\),
  \(\displaystyle {10^{0}}\),
  \(\displaystyle {10^{1}}\),
  \(\displaystyle {10^{2}}\),
  \(\displaystyle {10^{3}}\),
  \(\displaystyle {10^{4}}\),
  \(\displaystyle {10^{5}}\),
  \(\displaystyle {10^{6}}\),
  \(\displaystyle {10^{7}}\),
  \(\displaystyle {10^{8}}\)
},
y grid style={darkgray176},
ylabel={Number of samples},
ymin=0, ymax=418.560177280367,
ytick style={color=black},
ytick={0,50,100,150,200,250,300,350,400,450},
yticklabels={
  \(\displaystyle {0}\),
  \(\displaystyle {50}\),
  \(\displaystyle {100}\),
  \(\displaystyle {150}\),
  \(\displaystyle {200}\),
  \(\displaystyle {250}\),
  \(\displaystyle {300}\),
  \(\displaystyle {350}\),
  \(\displaystyle {400}\),
  \(\displaystyle {450}\)
},
separate axis lines,
axis x line=bottom,
axis y line=left,
x axis line style={-},
y axis line style={-}
]
\draw[draw=black,fill=yellowgreen13820138,fill opacity=0.6] (axis cs:9.28317766722556,0) rectangle (axis cs:20,2);
\draw[draw=black,fill=yellowgreen13820138,fill opacity=0.6] (axis cs:92.8317766722556,0) rectangle (axis cs:200,92);
\draw[draw=black,fill=yellowgreen13820138,fill opacity=0.6] (axis cs:928.317766722556,0) rectangle (axis cs:2000,216);
\draw[draw=black,fill=yellowgreen13820138,fill opacity=0.6] (axis cs:9283.17766722556,0) rectangle (axis cs:20000,121);
\draw[draw=black,fill=yellowgreen13820138,fill opacity=0.6] (axis cs:92831.7766722557,0) rectangle (axis cs:200000,8);
\draw[draw=black,fill=yellowgreen13820138,fill opacity=0.6] (axis cs:928317.766722557,0) rectangle (axis cs:2000000,0);
\draw[draw=black,fill=sandybrown25520258,fill opacity=0.6] (axis cs:4.30886938006377,0) rectangle (axis cs:9.28317766722556,11);
\draw[draw=black,fill=sandybrown25520258,fill opacity=0.6] (axis cs:43.0886938006377,0) rectangle (axis cs:92.8317766722556,120);
\draw[draw=black,fill=sandybrown25520258,fill opacity=0.6] (axis cs:430.886938006377,0) rectangle (axis cs:928.317766722556,276);
\draw[draw=black,fill=sandybrown25520258,fill opacity=0.6] (axis cs:4308.86938006377,0) rectangle (axis cs:9283.17766722556,101);
\draw[draw=black,fill=sandybrown25520258,fill opacity=0.6] (axis cs:43088.6938006377,0) rectangle (axis cs:92831.7766722557,4);
\draw[draw=black,fill=sandybrown25520258,fill opacity=0.6] (axis cs:430886.938006376,0) rectangle (axis cs:928317.766722557,0);
\draw[draw=black,fill=steelblue25130196,fill opacity=0.6] (axis cs:2,0) rectangle (axis cs:4.30886938006377,0);
\draw[draw=black,fill=steelblue25130196,fill opacity=0.6] (axis cs:20,0) rectangle (axis cs:43.0886938006377,13);
\draw[draw=black,fill=steelblue25130196,fill opacity=0.6] (axis cs:200,0) rectangle (axis cs:430.886938006377,293);
\draw[draw=black,fill=steelblue25130196,fill opacity=0.6] (axis cs:2000,0) rectangle (axis cs:4308.86938006377,280);
\draw[draw=black,fill=steelblue25130196,fill opacity=0.6] (axis cs:20000,0) rectangle (axis cs:43088.6938006377,123);
\draw[draw=black,fill=steelblue25130196,fill opacity=0.6] (axis cs:200000,0) rectangle (axis cs:430886.938006376,30);
\addplot [semithick, yellowgreen13820138]
table {%
2 1.98947071497049
2.13472148379802 1.98247811708357
2.2785179066944 1.94348248220454
2.43200056331951 1.87718164123201
2.59582192556351 1.79029434988615
2.77067841630718 1.69104355822957
2.9573133699932 1.58849684817434
3.1565201926238 1.4918277865969
3.36914573461814 1.409566973172
3.59609389086789 1.34890901193746
3.83832944329524 1.31513322735421
4.09688216224841 1.31118383701364
4.37285118417027 1.3374414288412
4.66740968414994 1.39170326289726
4.98180986322089 1.46937565843993
5.31738827160724 1.56386753343713
5.67557149054779 1.66715994577071
6.05788219685194 1.77051253906166
6.46594563596868 1.86525510727887
6.90149653108618 1.94360273059998
7.36638645763357 1.99942811092762
7.86259171453457 2.02892661801097
8.39222172567461 2.03111904026508
8.95752800729702 2.00815367223528
9.56091373944969 1.96539125626249
10.2049439821714 1.91128035356097
10.8923565798483 1.85705334856131
11.6260738000954 1.81629117980968
12.4092147566425 1.80441577165747
13.245108669034 1.83817224475983
14.1373090155131 1.93515907410921
15.0896086392536 2.11345533654943
16.1060558721594 2.391382367291
17.1909717447749 2.78742449160986
18.3489683554679 3.32032101353621
19.5849684769736 4.00932913494234
20.9042264836512 4.87464380267514
22.3123506884148 5.93794420376195
23.8153271842973 7.22301691085909
25.4195452919992 8.75638311965823
27.1318247216036 10.5678345554226
28.9594445639247 12.6907638872556
30.9101742347338 15.1621665117007
32.9923065034131 18.0221969600918
35.2146927464425 21.3131890035755
37.5867805755884 25.0780948366379
40.1186540007553 29.3583626121211
42.8210762982357 34.1913460484587
45.7055357665989 39.6074144683312
48.7842945647286 45.6269944324549
52.0704408396285 52.2578136694623
55.5779443655942 59.492625775517
59.3217159312824 67.3076664956425
63.3176707271358 75.662031427155
67.5827960026327 84.4980783971542
72.1352232809793 93.7428574241178
76.9943054382367 103.310470855332
82.1806989745551 113.105179088163
87.7164518272601 123.025003392903
93.6250970990929 132.965542501393
99.9317531000544 142.823714789803
106.663230128143 152.501160028439
113.848144442919 161.907078242219
121.517039916419 170.960341873344
129.70251787856 179.590785393184
138.43937570903 187.739649198451
147.764754764825 195.359228409043
157.718298272309 202.411848457269
168.34231985498 208.868353501353
179.681983413412 214.706344473594
191.785495122025 219.908433435635
204.704308358913 224.460782348212
218.493342439892 228.352162016442
233.211216086637 231.573699903353
248.920496621403 234.119388783665
265.68796594769 235.987313176518
283.584904447565 237.181433371926
302.687394002512 237.713665923633
323.076641425998 237.605931442856
344.839323682693 236.891817145992
368.067956361911 235.617527342558
392.861286971701 233.841866409734
419.324714725514 231.635105452324
447.570738606014 229.076711020934
477.719435610802 226.252045708338
509.898971213121 223.248271121087
544.246144207577 220.149781596528
580.906968257074 217.033563780976
620.037292613174 213.964907962024
661.803464648649 210.993889778522
706.383037018716 208.152995587534
753.965522457171 205.456183031886
804.753199416159 202.899553779301
858.961971974431 200.46367395168
916.822287669663 198.11741840282
978.580117156637 195.823050072658
1044.49799985593 193.542090653396
1114.85616003825 191.241411507618
1189.95369808911 188.898892517805
1270.10986201786 186.507977890955
1355.66540461662 184.080513913702
1446.98403203842 181.647388876699
1544.45394995254 179.256704679506
1648.4895138502 176.969476634676
1759.53299051588 174.853155440203
1878.05643815281 172.973557259706
2004.56371315499 171.386034072544
2139.59261205694 170.126878224998
2283.71715776673 169.206001872469
2437.5500398014 168.601847381446
2601.74521889837 168.259270763831
2777.00070707556 168.0908185208
2964.06153495824 167.981428107137
3163.72291898734 167.796174998191
3376.83364197322 167.390320842214
3604.29966136606 166.620638725628
3847.08796058202 165.356841684731
4106.23065975757 163.491938744673
4382.82940340729 160.950486404542
4678.06004363759 157.693968836503
4993.17763882512 153.722886540377
5329.52178900991 149.075510556951
5688.52233068454 143.823616194279
6071.70541518853 138.065802167177
6480.69999654785 131.919197037992
6917.24475634021 125.510440606117
7383.1954950243 118.966806105111
7880.53302115456 112.408216598748
8411.37157201915 105.940732117214
8977.96780149858 99.651873031036
9582.73037335293 93.6079288993823
10228.2302007201 87.8532045147231
10917.2113753545 82.4109929685827
11652.6028330666 77.2859486392286
12437.5308049565 72.4674637248606
13275.3321073701 67.9336278367819
14169.5683270783 63.6553652999967
15124.0409619789 59.6003913112763
16142.8075816888 55.7366972754403
17230.1990767243 52.0353586723005
18390.8380696001 48.4725472188501
19629.6585661128 45.030715020004
20951.9269303504 41.6989948840731
22363.2642725926 38.4729223167946
23869.670345278 35.3536269053693
25477.5490486207 32.3466620113913
27193.7356543041 29.4606424969145
29025.5258629836 26.7058436236091
30980.7068191229 24.0928852926767
33067.5902150147 21.6315905644849
35295.0476247104 19.3300720281713
37672.5482180718 17.1940690149695
40210.1990152672 15.2265357324229
42918.7878528423 13.4274656103916
45809.8292440159 11.7939288879538
48895.6133281597 10.3202959274489
52189.2581175515 8.99861506351261
55704.7652635086 7.81910912769076
59457.0795789686 6.77074889176186
63462.1525705562 5.84185590270497
67737.010250217 5.02068389078806
72299.8255146922 4.29592950216978
77169.9954005307 3.65713097871022
82368.2235430534 3.09492742840721
87916.6081898169 2.60116964077731
93838.7361427273 2.1688928745266
100159.783028167 1.79217911181737
106906.620321387 1.46594786703388
114107.929580152 1.18571895372796
121794.32437323 0.947387483026498
129998.4804221 0.747042171746825
138755.274509076 0.580845237812079
148101.932742407 0.444978607547678
158078.188808613 0.33564941746094
168726.452884813 0.249139627750623
180091.991929122 0.181880710662658
192223.122115537 0.130534585662745
205171.414231383 0.0920652748244483
218991.912910478 0.0637908173190713
233743.370634011 0.0434104817125315
249488.497493892 0.0290071857273185
266294.227780349 0.0190286182843388
284232.004527057 0.0122526220456159
303378.083223442 0.00774304344670199
323813.855985272 0.0048018160172932
345626.197561618 0.00292192326712866
368907.834649101 0.00174448252021142
393757.740083422 0.00102180888509476
420281.553583917 0.000587155888249397
448592.030839797 0.000330976599254169
478809.522847148 0.000183013884239225
511062.487534443 9.9265700358731e-05
545488.035851515 5.28117650068411e-05
582232.514643505 2.75593061474984e-05
621452.128787616 1.4105986524436e-05
663313.605237467 7.08156212847029e-06
707994.901797969 3.48689663216508e-06
755685.963643795 1.6839469853505e-06
806589.530797507 7.97614133398465e-07
860921.999999999 3.7053460485609e-07
};
\addplot [semithick, sandybrown25520258]
table {%
2 9.6888342645965
2.13472148379802 9.60121620821369
2.2785179066944 9.33862237445252
2.43200056331951 8.91688134696051
2.59582192556351 8.36060135931987
2.77067841630718 7.70121950131523
2.9573133699932 6.97461134456796
3.1565201926238 6.21855473217362
3.36914573461814 5.47033497576166
3.59609389086789 4.76473019871636
3.83832944329524 4.13253645007608
4.09688216224841 3.59969842916177
4.37285118417027 3.18702024964012
4.66740968414994 2.91035644662125
4.98180986322089 2.78113645140415
5.31738827160724 2.80706005445041
5.67557149054779 2.99281500932983
6.05788219685194 3.34070418395529
6.46594563596868 3.85111894743978
6.90149653108618 4.52284742720584
7.36638645763357 5.35325161575231
7.86259171453457 6.33837920729663
8.39222172567461 7.47309073396947
8.95752800729702 8.75127937563639
9.56091373944969 10.1662417059873
10.2049439821714 11.7112265236571
10.8923565798483 13.3801509214923
11.6260738000954 15.1684335637702
12.4092147566425 17.0738605473123
13.245108669034 19.0973746727538
14.1373090155131 21.2436691415366
15.0896086392536 23.52147497876
16.1060558721594 25.9434591365555
17.1909717447749 28.5256956594306
18.3489683554679 31.286730446018
19.5849684769736 34.2463225110967
20.9042264836512 37.4240001158276
22.3123506884148 40.8376067942541
23.8153271842973 44.5020201092239
25.4195452919992 48.4281995898454
27.1318247216036 52.6226614463606
28.9594445639247 57.0873960082708
30.9101742347338 61.8201559665009
32.9923065034131 66.8149697134127
35.2146927464425 72.0626936108753
37.5867805755884 77.5514228514481
40.1186540007553 83.2666351760166
42.8210762982357 89.1910351261183
45.7055357665989 95.3041776475463
48.7842945647286 101.582050900535
52.0704408396285 107.99686114634
55.5779443655942 114.517266733833
59.3217159312824 121.109245592775
63.3176707271358 127.737659133928
67.5827960026327 134.36841777501
72.1352232809793 140.970992391798
76.9943054382367 147.52088785799
82.1806989745551 154.001630681334
87.7164518272601 160.405842102482
93.6250970990929 166.735074490293
99.9317531000544 172.998269118378
106.663230128143 179.208919486001
113.848144442919 185.381258516076
121.517039916419 191.52598908009
129.70251787856 197.646207596369
138.43937570903 203.734201773724
147.764754764825 209.769722347044
157.718298272309 215.720138949724
168.34231985498 221.542614563963
179.681983413412 227.188110192096
191.785495122025 232.606712026236
204.704308358913 237.753512196287
218.493342439892 242.594120933286
233.211216086637 247.108878055231
248.920496621403 251.294978070838
265.68796594769 255.166012064587
283.584904447565 258.748820551597
302.687394002512 262.077983533268
323.076641425998 265.188675315617
344.839323682693 268.108912858538
368.067956361911 270.852373357369
392.861286971701 273.412920787122
419.324714725514 275.761764534263
447.570738606014 277.847808620594
477.719435610802 279.601294150772
509.898971213121 280.940361030572
544.246144207577 281.779729791635
580.906968257074 282.040392516618
620.037292613174 281.659046332807
661.803464648649 280.596022582505
706.383037018716 278.840653476461
753.965522457171 276.413347469222
804.753199416159 273.36406947646
858.961971974431 269.767386676878
916.822287669663 265.714685705361
978.580117156637 261.304536157843
1044.49799985593 256.632420920278
1114.85616003825 251.781142385316
1189.95369808911 246.813130046471
1270.10986201786 241.765625703765
1355.66540461662 236.649336625903
1446.98403203842 231.45067437656
1544.45394995254 226.137203124106
1648.4895138502 220.665478558247
1759.53299051588 214.990135201593
1878.05643815281 209.072927847059
2004.56371315499 202.890477434162
2139.59261205694 196.439705548242
2283.71715776673 189.740324683493
2437.5500398014 182.834216898605
2601.74521889837 175.782000814353
2777.00070707556 168.65747672941
2964.06153495824 161.540889476497
3163.72291898734 154.512024788328
3376.83364197322 147.644058117673
3604.29966136606 140.998838407361
3847.08796058202 134.623970468962
4106.23065975757 128.551726104927
4382.82940340729 122.799529255291
4678.06004363759 117.371570556429
4993.17763882512 112.26103313021
5329.52178900991 107.452449037202
5688.52233068454 102.923827460221
6071.70541518853 98.6483609992702
6480.69999654785 94.5956826138232
6917.24475634021 90.7327771235377
7383.1954950243 87.0247259948955
7880.53302115456 83.4354768531582
8411.37157201915 79.9287887158872
8977.96780149858 76.4694290448971
9582.73037335293 73.0246117514033
10228.2302007201 69.5655867202058
10917.2113753545 66.069235839898
11652.6028330666 62.5195050368579
12437.5308049565 58.9085064951694
13275.3321073701 55.2371550738173
14169.5683270783 51.5152501780249
15124.0409619789 47.7609707006785
16142.8075816888 43.9998086177701
17230.1990767243 40.2630202435228
18390.8380696001 36.5857181949929
19629.6585661128 33.0047580529201
20951.9269303504 29.5565888190437
22363.2642725926 26.2752340504208
23869.670345278 23.1905511054208
25477.5490486207 20.3268813880752
27193.7356543041 17.7021590906685
29025.5258629836 15.3274957613966
30980.7068191229 13.2072100496366
33067.5902150147 11.3392329059248
35295.0476247104 9.7157934564988
37672.5482180718 8.32428223146371
40210.1990152672 7.14819577583651
42918.7878528423 6.16808634998152
45809.8292440159 5.36246677346366
48895.6133281597 4.7086470432229
52189.2581175515 4.18350039465024
55704.7652635086 3.76416812913718
59457.0795789686 3.42871363795177
63462.1525705562 3.1567282060221
67737.010250217 2.92987815824411
72299.8255146922 2.73236959710886
77169.9954005307 2.55129807090104
82368.2235430534 2.37684934705856
87916.6081898169 2.20232525812317
93838.7361427273 2.02398421985689
100159.783028167 1.84070642437094
106906.620321387 1.65351469866324
114107.929580152 1.464999329588
121794.32437323 1.27870543929934
129998.4804221 1.09854300183294
138755.274509076 0.928272434277073
148101.932742407 0.771104659637343
158078.188808613 0.629436537906629
168726.452884813 0.504723917822512
180091.991929122 0.397478286166455
192223.122115537 0.307361242983059
205171.414231383 0.233344837201112
218991.912910478 0.173905057614765
233743.370634011 0.127219493838354
249488.497493892 0.091346836281244
266294.227780349 0.0643738120995385
284232.004527057 0.0445228868670652
303378.083223442 0.0302205359799442
323813.855985272 0.0201305279328721
345626.197561618 0.0131593384101437
368907.834649101 0.00844174798192433
393757.740083422 0.0053142882230925
420281.553583917 0.0032829803282131
448592.030839797 0.00199020963701912
478809.522847148 0.00118395262025989
511062.487534443 0.000691150907583039
545488.035851515 0.000395925086596395
582232.514643505 0.000222563310637588
621452.128787616 0.000122770387679535
663313.605237467 6.64557352252555e-05
707994.901797969 3.52995793393491e-05
755685.963643795 1.83994302071151e-05
806589.530797507 9.41101904091398e-06
860921.999999999 4.7235236221209e-06
};
\addplot [semithick, steelblue25130196]
table {%
2 2.33405540851064e-08
2.13472148379802 5.04069286486075e-08
2.2785179066944 1.07062927973074e-07
2.43200056331951 2.23644845099474e-07
2.59582192556351 4.59463320900461e-07
2.77067841630718 9.28360152822436e-07
2.9573133699932 1.84483531205617e-06
3.1565201926238 3.6055894628421e-06
3.36914573461814 6.93066610850248e-06
3.59609389086789 1.31025737094188e-05
3.83832944329524 2.43626543889436e-05
4.09688216224841 4.45536194589487e-05
4.37285118417027 8.01376366074905e-05
4.66740968414994 0.000141772119589248
4.98180986322089 0.000246690551279367
5.31738827160724 0.000422210847049142
5.67557149054779 0.000710772567552267
6.05788219685194 0.00117697489849136
6.46594563596868 0.00191713125676339
6.90149653108618 0.00307184818507949
7.36638645763357 0.00484204461882082
7.86259171453457 0.00750861919601264
8.39222172567461 0.0114556188168972
8.95752800729702 0.0171962455650026
9.56091373944969 0.0254003710095537
10.2049439821714 0.0369214535003174
10.8923565798483 0.0528199680669555
11.6260738000954 0.0743798008719385
12.4092147566425 0.103113711134488
13.245108669034 0.140754119842115
14.1373090155131 0.189226322350874
15.0896086392536 0.25060284797993
16.1060558721594 0.327040090756941
17.1909717447749 0.42070133749144
18.3489683554679 0.533673571280987
19.5849684769736 0.667888424419121
20.9042264836512 0.825059801059489
22.3123506884148 1.0066514155071
23.8153271842973 1.2138863821603
25.4195452919992 1.44780792144086
27.1318247216036 1.70939546476207
28.9594445639247 1.99973459662
30.9101742347338 2.32023331944499
32.9923065034131 2.67287215180636
35.2146927464425 3.06047253580763
37.5867805755884 3.48696751209552
40.1186540007553 3.95766060740296
42.8210762982357 4.47946269085242
45.7055357665989 5.06110094751603
48.7842945647286 5.71329756358203
52.0704408396285 6.44891681970778
55.5779443655942 7.28307723172811
59.3217159312824 8.23322029132193
63.3176707271358 9.31912048042758
67.5827960026327 10.5628147783684
72.1352232809793 11.9884266165862
76.9943054382367 13.6218618152007
82.1806989745551 15.4903642348714
87.7164518272601 17.6219369214449
93.6250970990929 20.0446586780753
99.9317531000544 22.7859525874042
106.663230128143 25.871886860551
113.848144442919 29.3266037451561
121.517039916419 33.1719738577358
129.70251787856 37.4275577127845
138.43937570903 42.1109225893598
147.764754764825 47.2383136257076
157.718298272309 52.8256187931077
168.34231985498 58.8895063836151
179.681983413412 65.448560500602
191.785495122025 72.5242043608505
204.704308358913 80.1411909572088
218.493342439892 88.3274606836226
233.211216086637 97.1132166982009
248.920496621403 106.529147374507
265.68796594769 116.603823133566
283.584904447565 127.36040078076
302.687394002512 138.812868684599
323.076641425998 150.962146975182
344.839323682693 163.792406336304
368.067956361911 177.267978332295
392.861286971701 191.331195891072
419.324714725514 205.901426671808
447.570738606014 220.875452546386
477.719435610802 236.129218378768
509.898971213121 251.520839265304
544.246144207577 266.894635221147
580.906968257074 282.085872369579
620.037292613174 296.925842325049
661.803464648649 311.246912635232
706.383037018716 324.887229156446
753.965522457171 337.694836600283
804.753199416159 349.531090293343
858.961971974431 360.273340771361
916.822287669663 369.816963396715
978.580117156637 378.076861754594
1044.49799985593 384.988587350317
1114.85616003825 390.509189561223
1189.95369808911 394.617848774557
1270.10986201786 397.316269522209
1355.66540461662 398.628740267016
1446.98403203842 398.601722255852
1544.45394995254 397.302825586797
1648.4895138502 394.819070902659
1759.53299051588 391.254413528435
1878.05643815281 386.726607210908
2004.56371315499 381.36358445669
2139.59261205694 375.2996063652
2283.71715776673 368.671468037994
2437.5500398014 361.61502669017
2601.74521889837 354.262250970556
2777.00070707556 346.738885825528
2964.06153495824 339.162710185948
3163.72291898734 331.642260868626
3376.83364197322 324.275828743304
3604.29966136606 317.150517647239
3847.08796058202 310.341196187413
4106.23065975757 303.909258257776
4382.82940340729 297.901220009891
4678.06004363759 292.347293223481
4993.17763882512 287.260161356489
5329.52178900991 282.634224497274
5688.52233068454 278.445562713829
6071.70541518853 274.652796071559
6480.69999654785 271.198907892236
6917.24475634021 268.013968247329
7383.1954950243 265.018573245827
7880.53302115456 262.127726032641
8411.37157201915 259.254843674138
8977.96780149858 256.315585560442
9582.73037335293 253.231257327221
10228.2302007201 249.931633704778
10917.2113753545 246.357142730632
11652.6028330666 242.460440697972
12437.5308049565 238.207465085984
13275.3321073701 233.578072983087
14169.5683270783 228.566356514287
15124.0409619789 223.180684735252
16142.8075816888 217.443469307373
17230.1990767243 211.390606320569
18390.8380696001 205.070523073314
19629.6585661128 198.542764117705
20951.9269303504 191.876085020039
22363.2642725926 185.146077183455
23869.670345278 178.432409822336
25477.5490486207 171.815831421041
27193.7356543041 165.375110645611
29025.5258629836 159.184108657299
30980.7068191229 153.309160382088
33067.5902150147 147.806906640526
35295.0476247104 142.722671115881
37672.5482180718 138.089425971899
40210.1990152672 133.927345846106
42918.7878528423 130.243916595362
45809.8292440159 127.034542914848
48895.6133281597 124.283584634093
52189.2581175515 121.965740307559
55704.7652635086 120.0476845004
59457.0795789686 118.489850373977
63462.1525705562 117.248233860483
67737.010250217 116.276085228507
72299.8255146922 115.525355234449
77169.9954005307 114.947782694218
82368.2235430534 114.495551373821
87916.6081898169 114.121504731432
93838.7361427273 113.778979773744
100159.783028167 113.421393967245
106906.620321387 113.001777277319
114107.929580152 112.472471348675
121794.32437323 111.785210073419
129998.4804221 110.891747489103
138755.274509076 109.745115147957
148101.932742407 108.301484188755
158078.188808613 106.522494925154
168726.452884813 104.377818352315
180091.991929122 101.84764725111
192223.122115537 98.9247920015418
205171.414231383 95.616082714806
218991.912910478 91.94285137343
233743.370634011 87.9403742245233
249488.497493892 83.6562791297853
266294.227780349 79.1480459553486
284232.004527057 74.4798323189339
303378.083223442 69.7189279108232
323813.855985272 64.9321699323417
345626.197561618 60.1826385608311
368907.834649101 55.5268998600146
393757.740083422 51.0129843995563
420281.553583917 46.6791961826467
448592.030839797 42.553752050664
478809.522847148 38.6551686908777
511062.487534443 34.9932516424175
545488.035851515 31.5705031078341
582232.514643505 28.3837536023444
621452.128787616 25.4258336398472
663313.605237467 22.6871304303032
707994.901797969 20.1569144056285
755685.963643795 17.8243647253023
806589.530797507 15.679266030777
860921.999999999 13.7123863174258
};
\end{axis}

\end{tikzpicture}

%% file: figures/num_instances.pgf
\begin{tikzpicture}

\definecolor{darkgray176}{RGB}{176,176,176}
\definecolor{sandybrown23019082}{RGB}{230,190,82}
\definecolor{steelblue46125174}{RGB}{46,125,174}
\definecolor{yellowgreen13318058}{RGB}{133,180,58}

\begin{axis}[
tick align=outside,
tick pos=left,
x grid style={darkgray176},
xlabel={Instance per Scene},
xmin=-0.5, xmax=5.5,
xtick style={color=black},
xtick={0,1,2,3,4,5},
xticklabels={2,3,4,5,6,7+},
y grid style={darkgray176},
ylabel={Number of samples},
ymin=0, ymax=93.45,
ytick style={color=black},
ytick={0,20,40,60,80,100},
yticklabels={
  \(\displaystyle {0}\),
  \(\displaystyle {20}\),
  \(\displaystyle {40}\),
  \(\displaystyle {60}\),
  \(\displaystyle {80}\),
  \(\displaystyle {100}\)
},
separate axis lines,
axis x line=bottom,
axis y line=left,
x axis line style={-},
y axis line style={-}
]
\draw[draw=black,fill=steelblue46125174,opacity=0.6] (axis cs:-0.45,0) rectangle (axis cs:-0.15,15);
\addlegendimage{ybar,ybar legend,draw=black,fill=steelblue46125174,opacity=0.6}
\addlegendentry{roadanomaly}

\draw[draw=black,fill=steelblue46125174,opacity=0.6] (axis cs:0.55,0) rectangle (axis cs:0.85,12);
\draw[draw=black,fill=steelblue46125174,opacity=0.6] (axis cs:1.55,0) rectangle (axis cs:1.85,10);
\draw[draw=black,fill=steelblue46125174,opacity=0.6] (axis cs:2.55,0) rectangle (axis cs:2.85,3);
\draw[draw=black,fill=steelblue46125174,opacity=0.6] (axis cs:3.55,0) rectangle (axis cs:3.85,1);
\draw[draw=black,fill=steelblue46125174,opacity=0.6] (axis cs:4.55,0) rectangle (axis cs:4.85,18);
\draw[draw=black,fill=sandybrown23019082,opacity=0.6] (axis cs:-0.15,0) rectangle (axis cs:0.15,74);
\addlegendimage{ybar,ybar legend,draw=black,fill=sandybrown23019082,opacity=0.6}
\addlegendentry{roadobstacle}

\draw[draw=black,fill=sandybrown23019082,opacity=0.6] (axis cs:0.85,0) rectangle (axis cs:1.15,12);
\draw[draw=black,fill=sandybrown23019082,opacity=0.6] (axis cs:1.85,0) rectangle (axis cs:2.15,0);
\draw[draw=black,fill=sandybrown23019082,opacity=0.6] (axis cs:2.85,0) rectangle (axis cs:3.15,0);
\draw[draw=black,fill=sandybrown23019082,opacity=0.6] (axis cs:3.85,0) rectangle (axis cs:4.15,1);
\draw[draw=black,fill=sandybrown23019082,opacity=0.6] (axis cs:4.85,0) rectangle (axis cs:5.15,0);
\draw[draw=black,fill=yellowgreen13318058,opacity=0.6] (axis cs:0.15,0) rectangle (axis cs:0.45,89);
\addlegendimage{ybar,ybar legend,draw=black,fill=yellowgreen13318058,opacity=0.6}
\addlegendentry{fishyscapes}

\draw[draw=black,fill=yellowgreen13318058,opacity=0.6] (axis cs:1.15,0) rectangle (axis cs:1.45,38);
\draw[draw=black,fill=yellowgreen13318058,opacity=0.6] (axis cs:2.15,0) rectangle (axis cs:2.45,1);
\draw[draw=black,fill=yellowgreen13318058,opacity=0.6] (axis cs:3.15,0) rectangle (axis cs:3.45,0);
\draw[draw=black,fill=yellowgreen13318058,opacity=0.6] (axis cs:4.15,0) rectangle (axis cs:4.45,0);
\draw[draw=black,fill=yellowgreen13318058,opacity=0.6] (axis cs:5.15,0) rectangle (axis cs:5.45,0);
\end{axis}

\end{tikzpicture}

%% file: sections/4_results.tex
\input{tables/main_table}
\input{tables/detection_table_per_dataset}

\section{Experiments}
In this section we present quantitative results obtained by U3HS, Mask2Anomaly and UGainS on OoDIS' new annotations provided for the FS Lost \& Found, RoadAnomaly21 and RoadObstacle21.
The results for all three methods were computed using the original code repositories provided by the respectively publications, if available.
To ensure correctness, we contacted authors of the original works, and asked them for a submission to the benchmark.
We received a submission for Mask2Anomaly by the authors.
In case of U3HS the code was not available.
We worked closely with the authors to reimplement it and submit the results to the benchmark.
During this process, we kept the test sets of OoDIS private.
The authors used the validation set for debugging and parameter tuning.
In all experiments presented, we stick to default hyperparameters of the methods.
We used low score thresholds to obtain rather high numbers of predictions. The PPF counts provide a measure for this.

In this section, we first present a comparative study of the three methods on the three datasets for both tasks, anomaly instance segmentation and anomalous object detection, focusing on the popular metrics AP and AP50.
Thereafter we proceed with a more detailed analysis, providing results for further evaluation metrics.
This is complemented with a choice of qualitative results.

\parag{Anomaly Instance Segmentation.}
In 
\cref{tab:benchmark} we provide anomaly instance segmentation results for all three methods and all three datasets in terms of AP and AP50.
The three methods vary greatly in performance.
In terms of AP, none of them perform strongly, demonstrating that there is a room for further method development.
The AP50 scores for all of the data subsets are below 47\%, indicating that accurate localization on all three datasets is challenging.
A closer comparison of results on the different datasets reveals that RoadAnomaly21 is particularly challenging.
This can be explained as follows.
In RoadObstacle21, the number of connected components in the annotations increased from 388 for the semantic segmentation version to 557 instances for instance segmentation.
Most of the scenes contain one to two objects.
However, in RoadAnomaly21 this count considerably increased from 262 to 739.
The number of objects per scenes varies more strongly, with extreme cases where a flock of sheep consists of several dozens of instances.
These cases are particularly hard to segment accurately.
The segment sizes a summarized in \cref{fig:mask-distribution}.

\parag{Anomalous Object Detection.} In \cref{tab:detections_per_dataset}, an evaluation analogously to the previous paragraph on anomaly instance segmentation is provided for three methods over three datasets in terms of AP and AP50.
Compared to the results for anomalous instance segmentation, two key differences can be observed.
On the one hand, while the ranking of the methods is the same, a pronounced reduction in the AP and AP50 scores is observed. 
Smaller deviations of an instance segmentation, e.g.\ at the border of a component, are not punished by the segmentation IoU as much as for the object detection IoU.
In the latter metric, only slight deviations can have a strong effect.
On the other hand, while for anomalous instance segmentation the results on RoadAnomaly21 were considerably worse than on the other datasets, this is not the case anymore for anomalous object detection.
Here, the lowest scores are obtained on RoadObstacle21.
The reason for this might be due to the objects in RoadObstacle21 being smaller than in RoadAnomaly21.
Accurate localization becomes increasingly challenging when considering smaller objects.
In summary, the task of accurately localizing anomalous objects can be deemed a challenging task.
Note that for practical purposed such as automated driving or robotic control, localization accuracy matters.

\input{tables/detection_table}

\input{figures/qualitative}

\parag{Additional Evaluation Metrics.} In \cref{tab:detections_summarized}, we provide additional evaluation metrics for anomalous object detection.
All performance metrics are averaged over the three datasets.
The AR1, A10 and AR100 scores additionally indicate that it is particularly challenging for all the tested methods to detect the anomalous objects present in the data.
The majority of objects remain overlooked, also for the most strongly performing methods, namely UGainS.
Noteworthily, the small difference between AR10 and AR100 shows that all methods can hardly deal with the very crowded scenes of the dataset such as flocks of sheep on the roads.

\input{tables/size_table}

\parag{Evaluation for Different Object Sizes.}
In \cref{tab:size} we present instance segmentation results of the three methods, averaged over all three datasets for three different size ranges. This evaluation reveals that small objects are particularly difficult to find for all three methods. From a method perspective, it is noteworthy that UGainS and Mask2Anomaly are on par in medium sized object from 1.000--10.000 pixels, and Mask2Anomaly is even slightly superior for large objects. However, it gets clearly outperformed on the small objects below 1.000 pixels, which constitute the most challenging cases of the benchmark. Note that all mean values in the benchmark are computed as weighted averages where the weighting treats all images across all datasets uniformly. Hence, the size-related results do not average out to the provided mean results.

\parag{Qualitative Results.} \Cref{fig:qualitative} provides a qualitative comparison of the three methods on all three datasets.
Across all three datasets it can be observed that UH3S is able to spot the anomalous objects, however it is unable to provide accurate segmentation masks, which leads to the low numbers on the benchmarks.
Mask2Anomaly already provides strong segmentation performance, however, it can be seen from the qualitative examples that some instances are overlooked or multiple instances are merged into a joint segment.
UGainS is able to detect and accurately segment many of anomalous objects.
However, an overproduction of false positives is observed.
In summary, we conclude that there is still plenty of room do develop stronger anomaly instance segmentation and anomalous object detection methods.

%% file: tables/main_table.tex
\begin{table*}[ht]
    \setlength{\tabcolsep}{3pt}
    \caption{
    Evaluation of three existing anomaly segmentation methods for three different datasets in terms of AP and AP50. Higher scores indicate stronger performance.
    }
    \begin{tabularx}{\linewidth}{lccp{1pt}YYp{1pt}YYp{1pt}YYp{10pt}YY}
    \toprule
    \multirow{2}{*}[-0.5ex]{Method} & \multirow{2}{*}{\parbox{1cm}{\centering OOD Data}} & \multirow{2}{*}{\parbox{1.3cm}{\centering Extra Network}} && \multicolumn{2}{c}{\small{FishyScapes}} && \multicolumn{2}{c}{\small{RoadAnomaly21}} && \multicolumn{2}{c}{\small{RoadObstacle21}} && \multicolumn{2}{c}{\small{Mean}} \\
    \cmidrule{5-6} \cmidrule{8-9} \cmidrule{11-12} \cmidrule{14-15}
    && && \small{AP} & \small{AP50} && \small{AP} & \small{AP50} && \small{AP} & \small{AP50} && \small{AP} & \small{AP50} \\
    \midrule
    UGainS~\cite{nekrasov2023ugains} & \cmark & \cmark && 27.14 & 45.82 && 11.42 & 19.15 && 27.22 & 46.54 && 25.19 & 42.81 \\
    Mask2Anomaly~\cite{rai2023mask2anomaly} & \cmark & \xmark && 11.73 & 23.64 && \phantom{0}4.78 & \phantom{0}9.03 && 17.23 & 28.44 && 13.73 & 24.30 \\
    U3HS~\cite{gasperini2023u3hs} & \xmark & \xmark && \phantom{0}0.19 & \phantom{0}0.73 && \phantom{0}0.00 & \phantom{0}0.00 && \phantom{0}0.22 & \phantom{0}0.62 && \phantom{0}0.19 & \phantom{0}0.58 \\
    \bottomrule
    \end{tabularx}
    \label{tab:benchmark}
\end{table*}

%% file: tables/detection_table_per_dataset.tex
\begin{table*}[ht]
    \setlength{\tabcolsep}{3pt}
    \caption{
    Evaluation of the three methods from Table \ref{tab:benchmark} w.r.t.\ anomalous object detection. The detection results are split over three datasets.
    }
    \begin{tabularx}{\linewidth}{lccp{1pt}YYp{1pt}YYp{1pt}YYp{10pt}YY}
    \toprule
    \multirow{2}{*}[-0.5ex]{Method} & \multirow{2}{*}{\parbox{1cm}{\centering OOD Data}} & \multirow{2}{*}{\parbox{1.3cm}{\centering Extra Network}} && \multicolumn{2}{c}{\small{FishyScapes}} && \multicolumn{2}{c}{\small{RoadAnomaly21}} && \multicolumn{2}{c}{\small{RoadObstacle21}} && \multicolumn{2}{c}{\small{Mean}} \\
    \cmidrule{5-6} \cmidrule{8-9} \cmidrule{11-12} \cmidrule{14-15}
    && && \small{AP} & \small{AP50} && \small{AP} & \small{AP50} && \small{AP} & \small{AP50} && \small{AP} & \small{AP50} \\
    \midrule
    UGainS~\cite{nekrasov2023ugains} & \cmark & \cmark && 15.55 & 24.27 && 11.69 & 17.48 && \phantom{0}8.05 & 11.54 && 11.14 & 16.75 \\
    Mask2Anomaly~\cite{rai2023mask2anomaly} & \cmark & \xmark && \phantom{0}1.01 & \phantom{0}2.33 && \phantom{0}1.43 & \phantom{0}2.91 && \phantom{0}1.34 & \phantom{0}1.99 && \phantom{0}1.24 & \phantom{0}2.23 \\
    U3HS~\cite{gasperini2023u3hs} & \xmark & \xmark && \phantom{0}0.31 & \phantom{0}0.75 && \phantom{0}0.04 & \phantom{0}0.17 && \phantom{0}0.09 & \phantom{0}0.22 && \phantom{0}0.16 & \phantom{0}0.40 \\
    \bottomrule
    \end{tabularx}
    \label{tab:detections_per_dataset}
\end{table*}

%% file: tables/detection_table.tex
\begin{table*}[ht]
    \setlength{\tabcolsep}{3pt}
    \caption{
    Evaluation of three existing anomaly segmentation methods on the detection benchmark.
    The numbers are computed as weighted averages over all three datasets, where the weighting takes the datasets' sizes into account.
    }
    \begin{tabularx}{\linewidth}{lccp{1pt}YYYYYY}
    \toprule
    \multirow{2}{*}[-0.5ex]{Method} & \multirow{2}{*}{\parbox{1cm}{\centering OOD Data}} & \multirow{2}{*}{\parbox{1.3cm}{\centering Extra Network}} && \multicolumn{6}{c}{} \\
    && && \small{AP}$\uparrow$ & \small{AP50}$\uparrow$ & \small{AR1}$\uparrow$ & \small{AR10}$\uparrow$ & \small{AR100}$\uparrow$ & \small{PPF}$\downarrow$ \\
    \midrule
    UGainS~\cite{nekrasov2023ugains} & \cmark & \cmark && 11.14 & 16.75 & 14.98 & 39.07 & 41.45 & 12.36 \\
    Mask2Anomaly~\cite{rai2023mask2anomaly} & \cmark & \xmark && \phantom{0}1.24 & \phantom{0}2.23 & \phantom{0}0.09 & 18.87 & 19.45 & \phantom{0}9.74 \\
    U3HS~\cite{gasperini2023u3hs} & \xmark & \xmark && \phantom{0}0.16 & \phantom{0}0.40 & \phantom{0}1.08 & \phantom{0}1.82 & \phantom{0}1.83 & \phantom{0}3.80 \\
    \bottomrule
    \end{tabularx}
    \label{tab:detections_summarized}
\end{table*}

%% file: figures/qualitative.tex
\begin{figure*}
    \centering
    \scalebox{1.0}{
    \begin{tikzpicture}

        \node at (2, 7) {\small Label};
        \node at (6.1, 7) {\small U3HS};
        \node at (10.2, 7) {\small Mask2Anomaly};
        \node at (14.3, 7) {\small UGainS};

        \node[anchor=east, rotate=90, align=center, text width=2cm] at (-0.65, 2.125) {\small Fishyscapes L\&F};
        \node[anchor=east, rotate=90, align=center, text width=2cm] at (-0.65, 4.44) {\small Road Anomaly21};
        \node[anchor=east, rotate=90, align=center, text width=2cm] at (-0.65, 6.8) {\small Road Obstacle21};

        \node at (2,1) {\includegraphics[width=4cm]{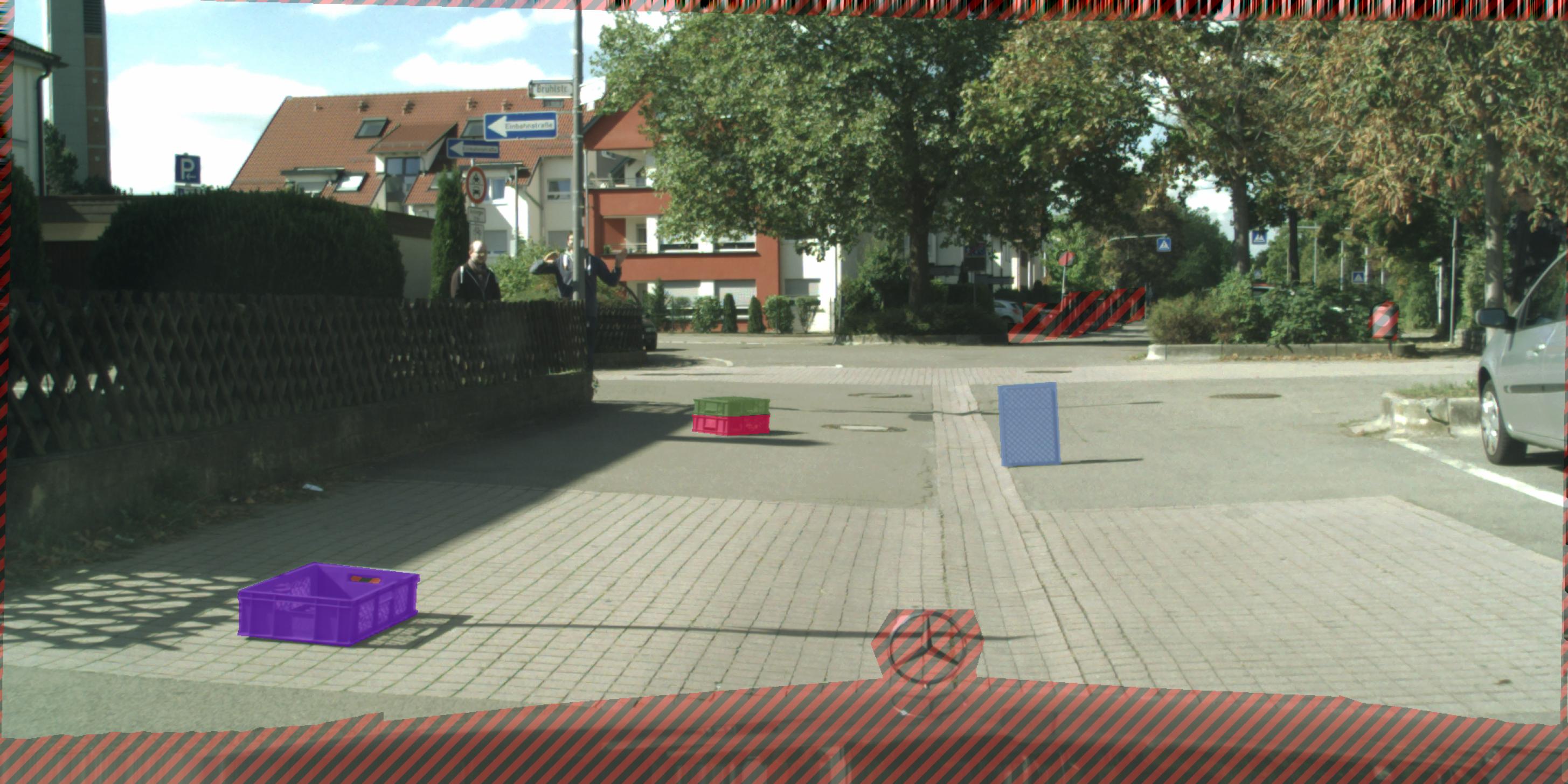}};
        \node at (2,3.25) {\includegraphics[width=4cm]{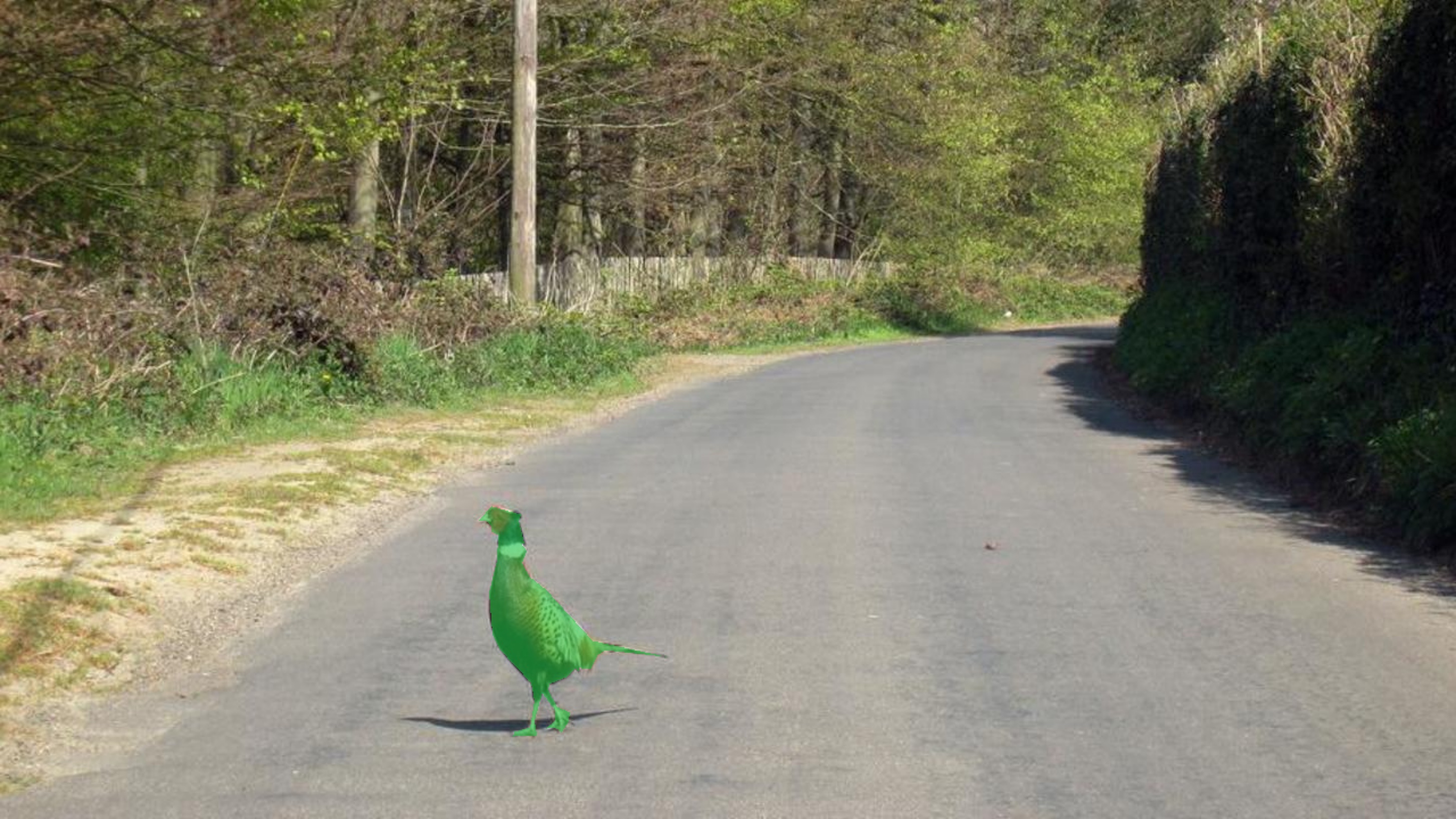}};
        \node at (2,5.63) {\includegraphics[width=4cm]{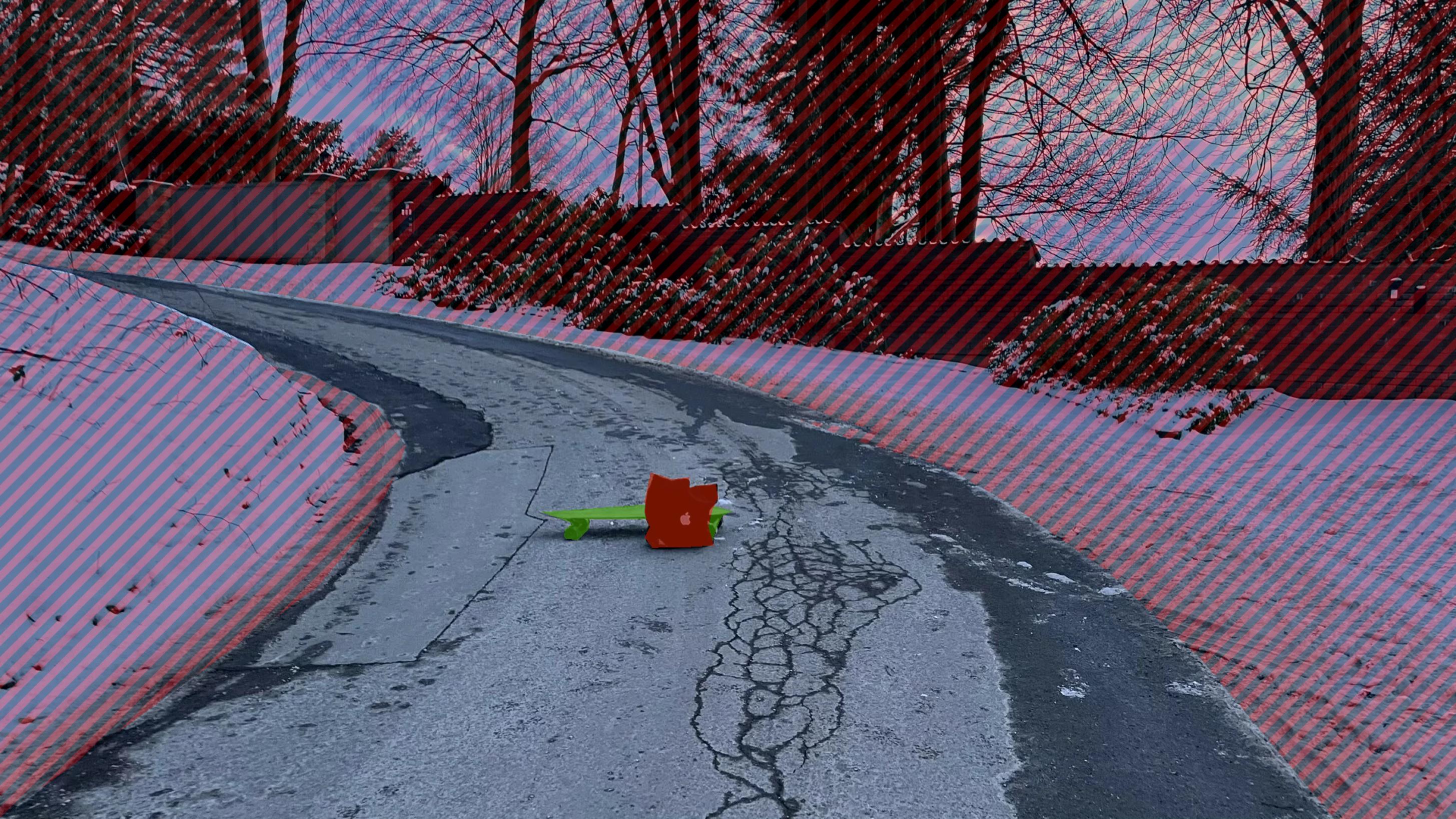}};

        \node at (6.1,1) {\includegraphics[width=4cm]{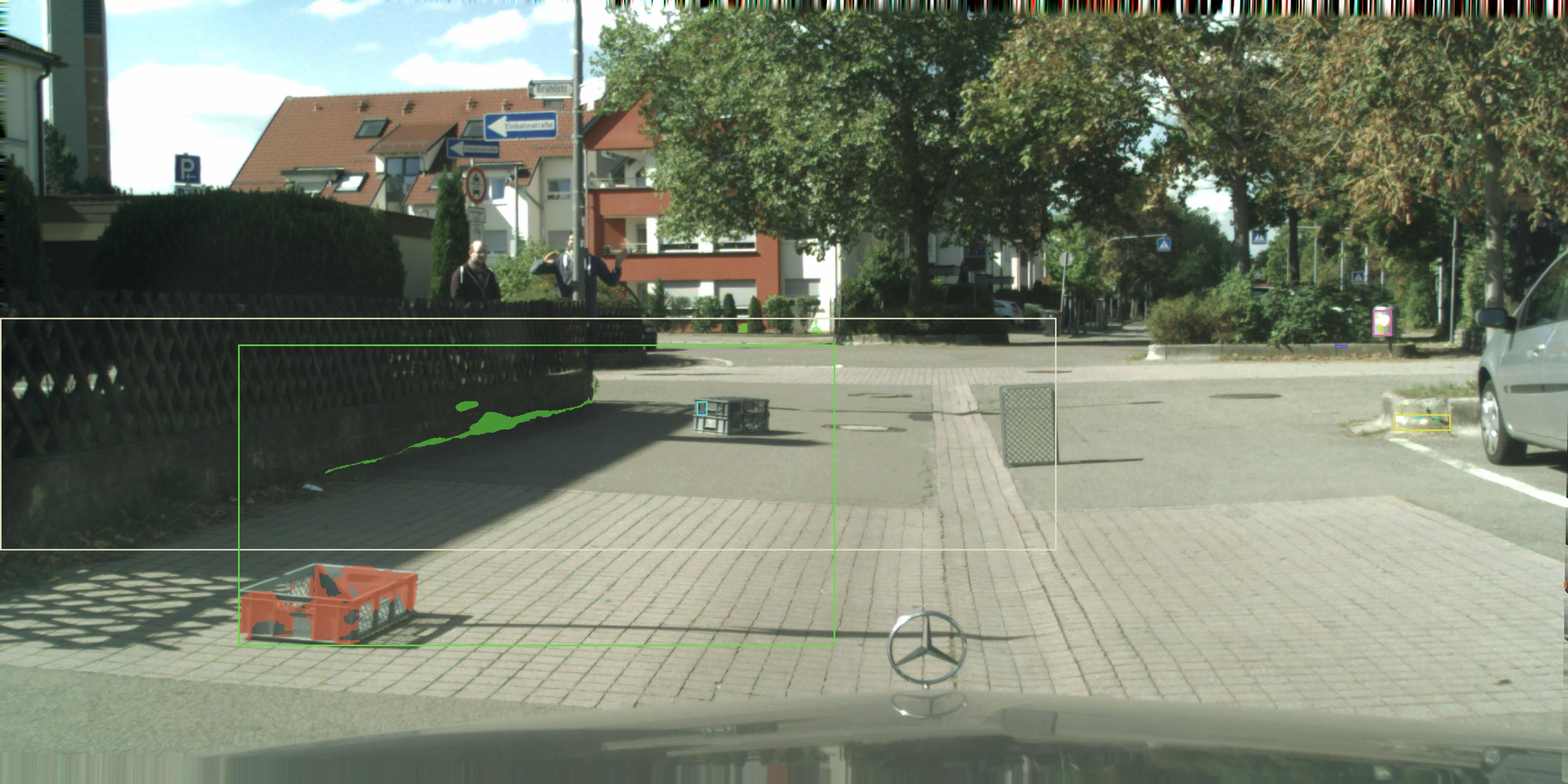}};
        \node at (6.1,3.25) {\includegraphics[width=4cm]{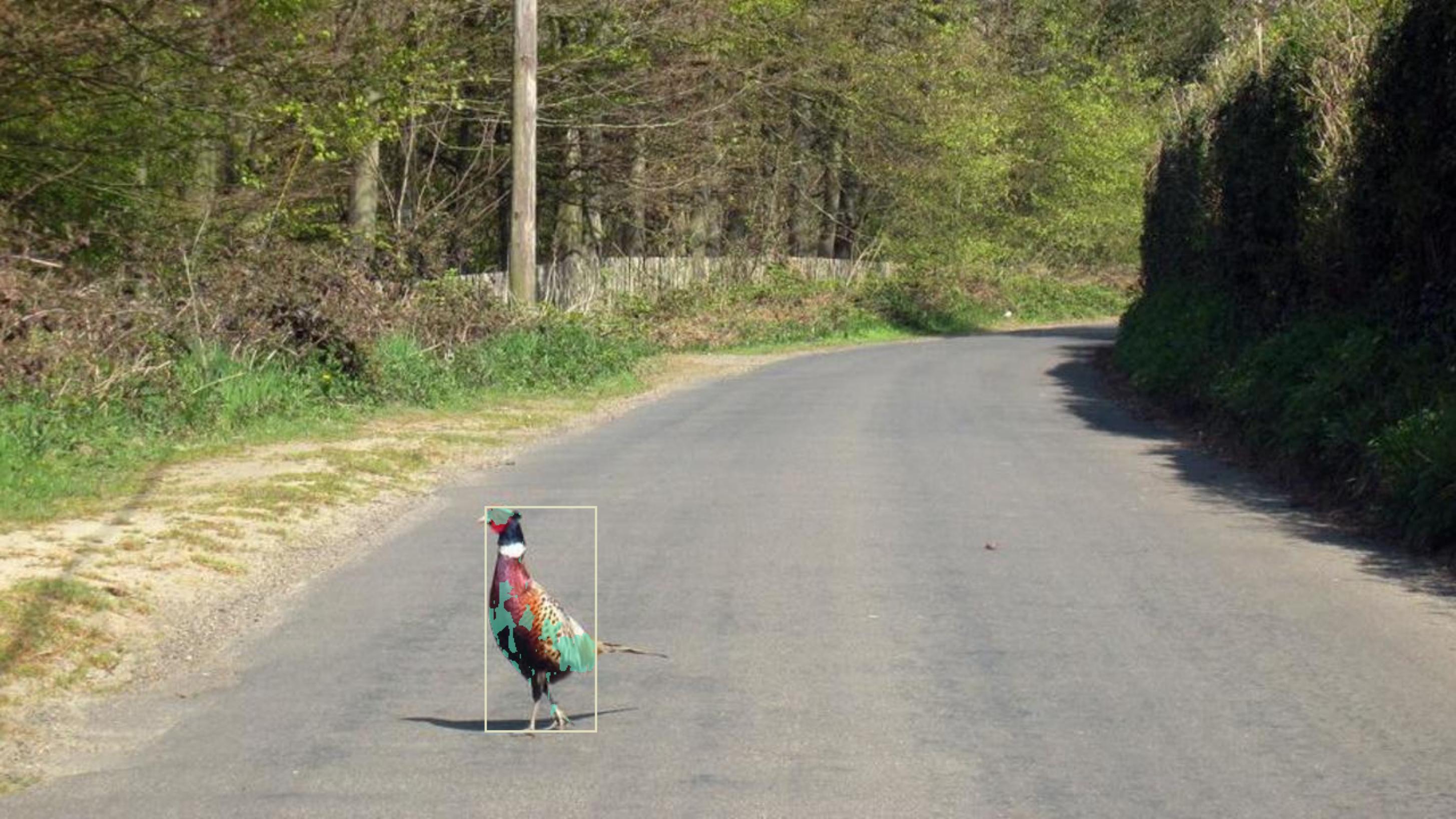}};
        \node at (6.1,5.63) {\includegraphics[width=4cm]{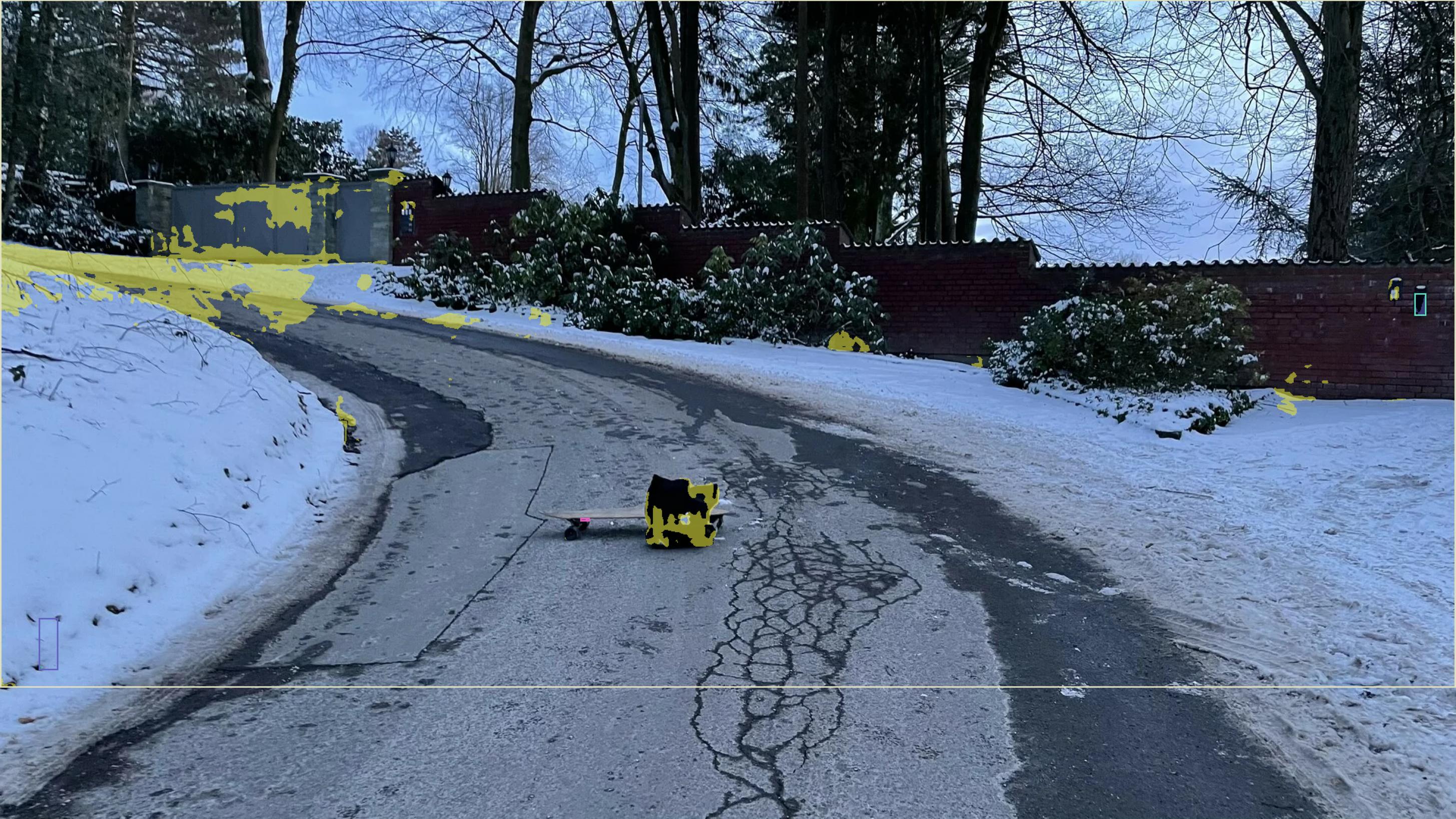}};

        \node at (10.2,1) {\includegraphics[width=4cm]{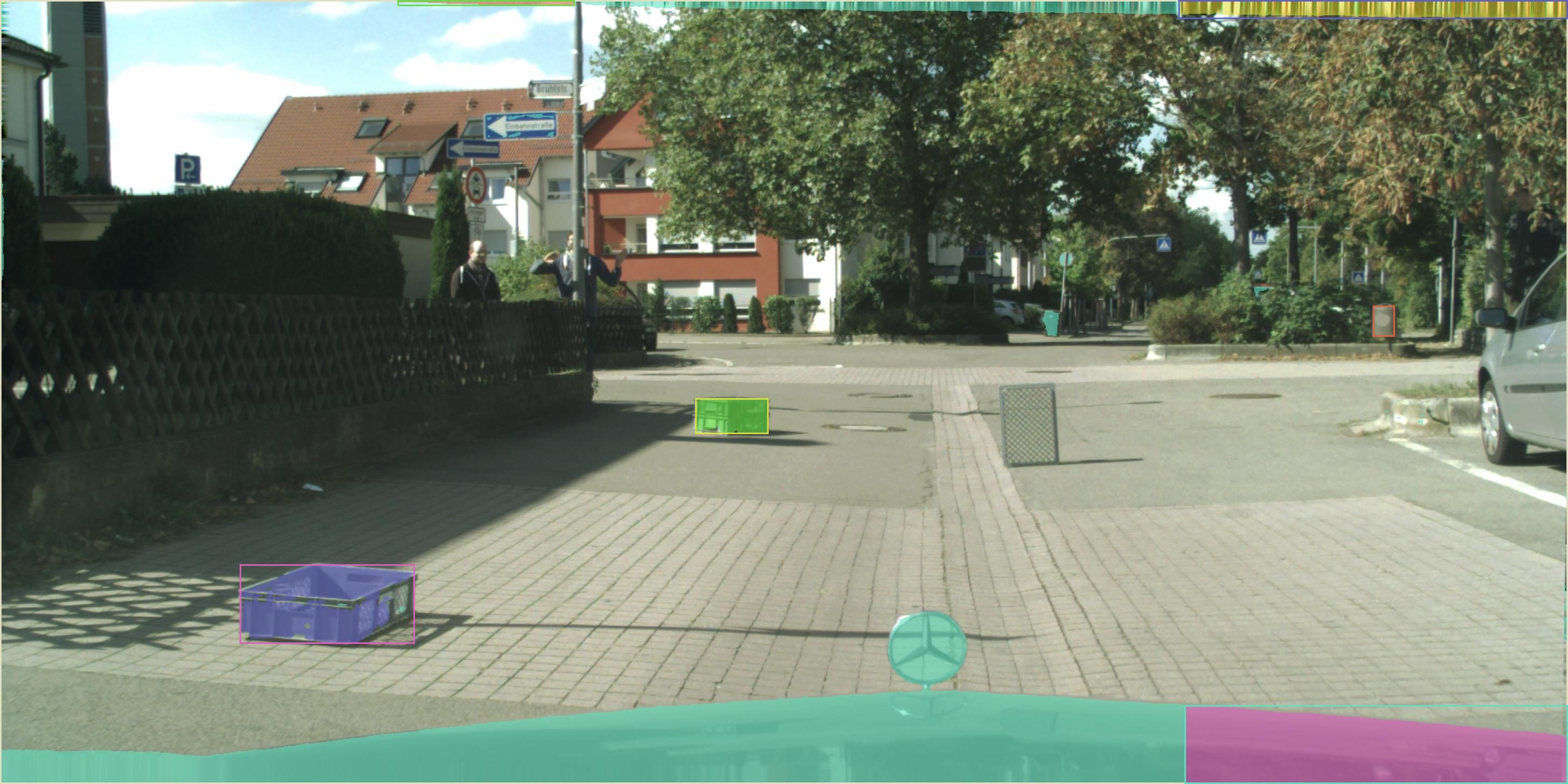}};
        \node at (10.2,3.25) {\includegraphics[width=4cm]{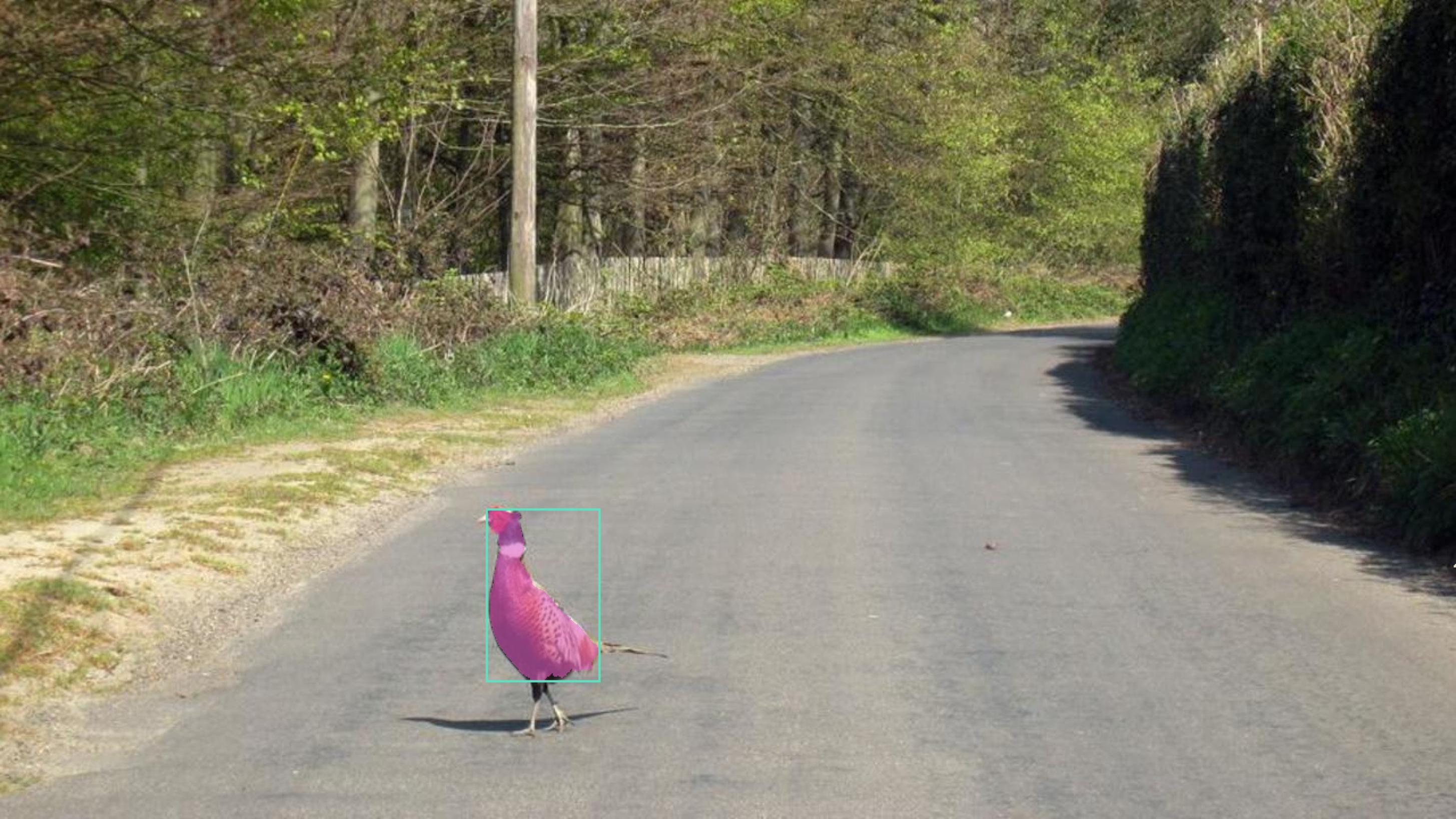}};
        \node at (10.2,5.63) {\includegraphics[width=4cm]{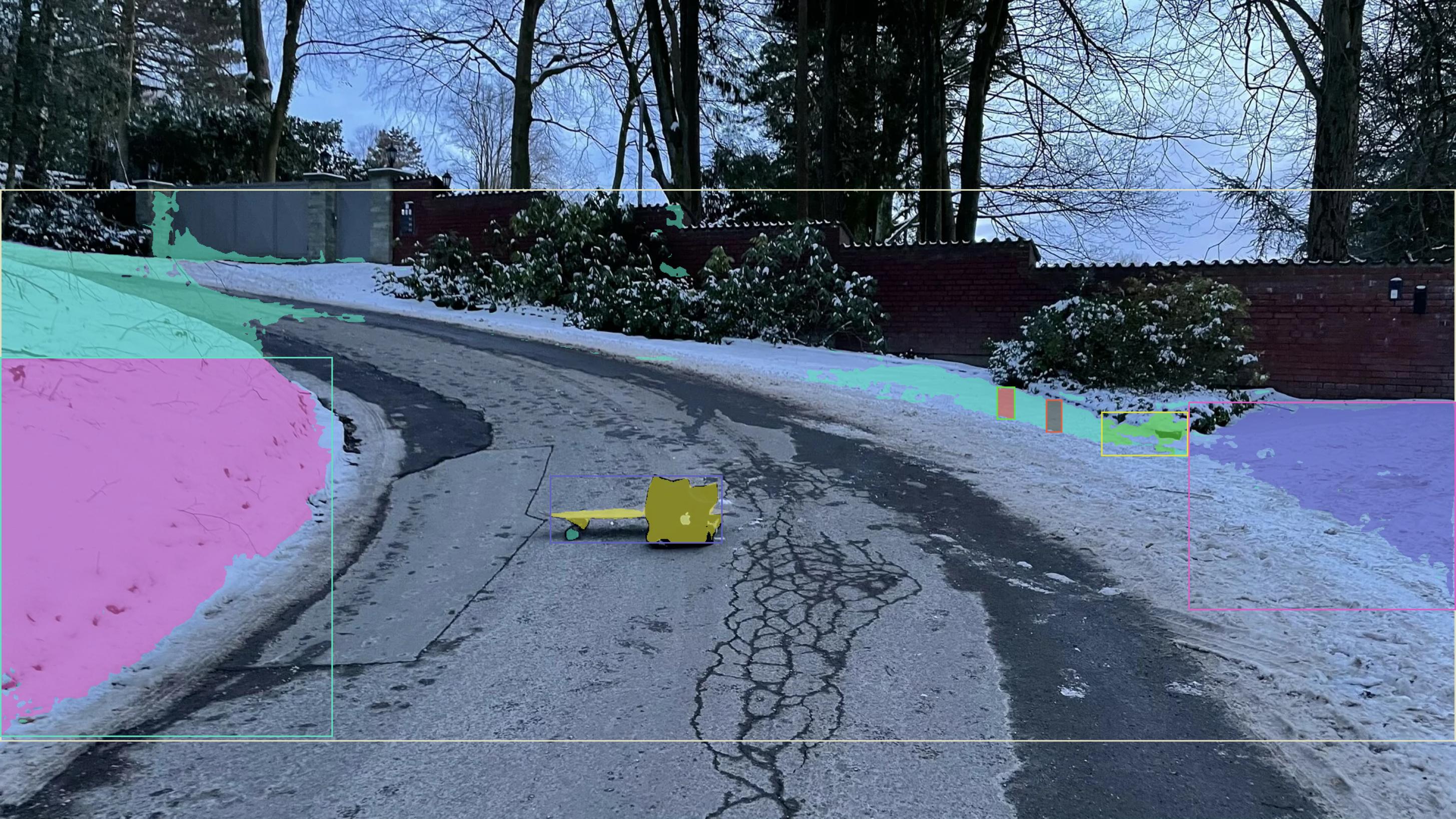}};

        \node at (14.3,1) {\includegraphics[width=4cm]{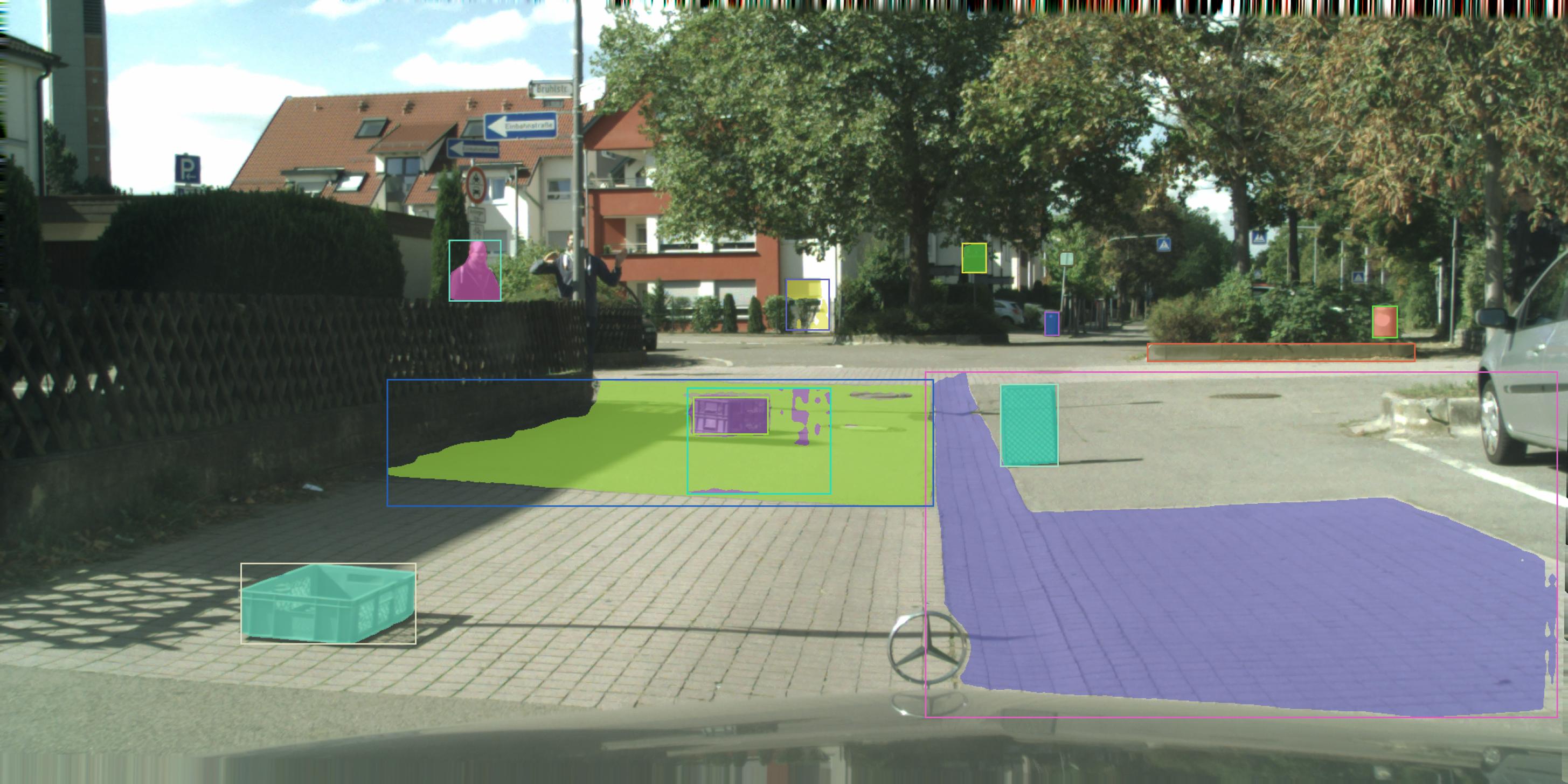}};
        \node at (14.3,3.25) {\includegraphics[width=4cm]{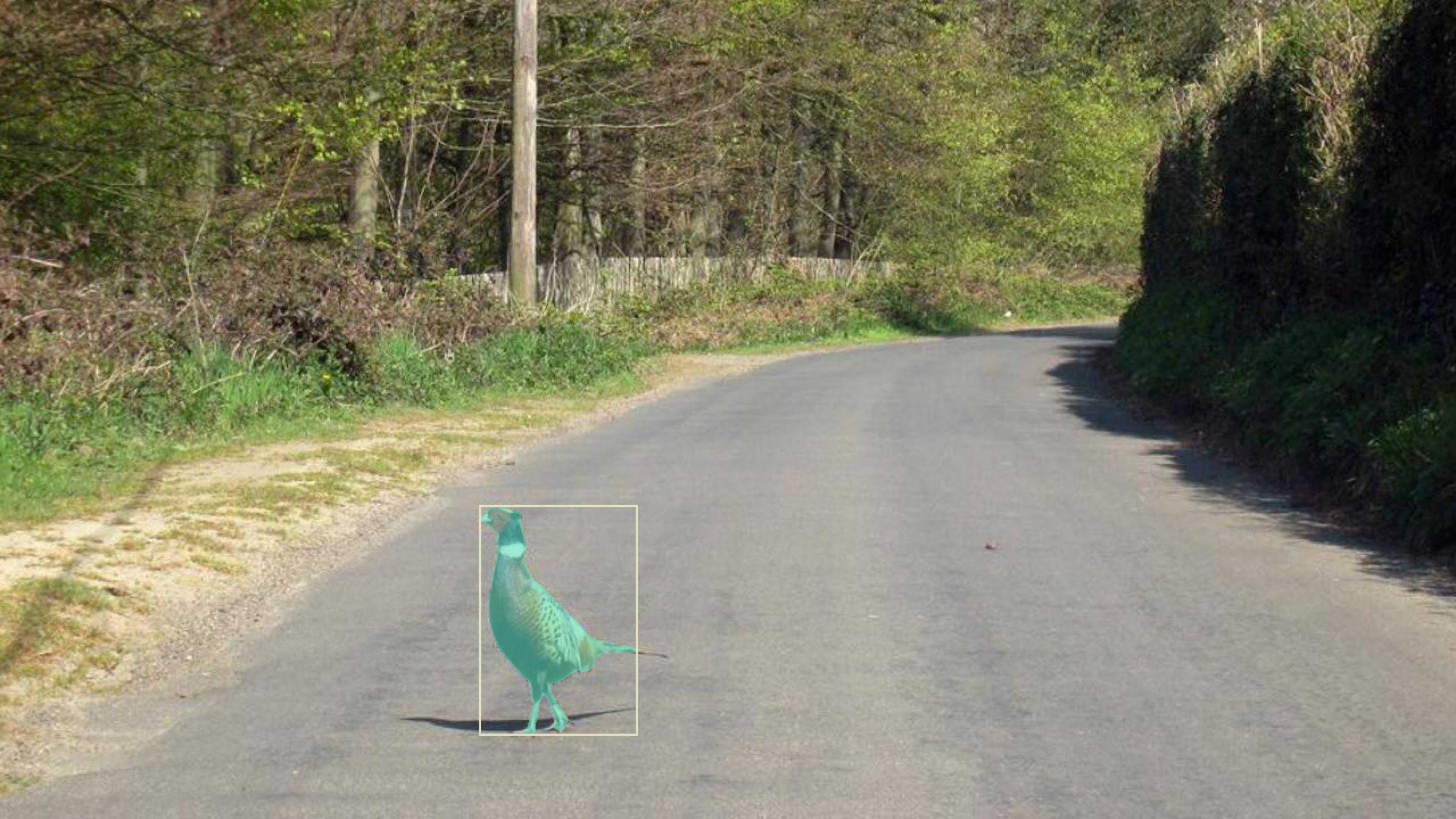}};
        \node at (14.3,5.63) {\includegraphics[width=4cm]{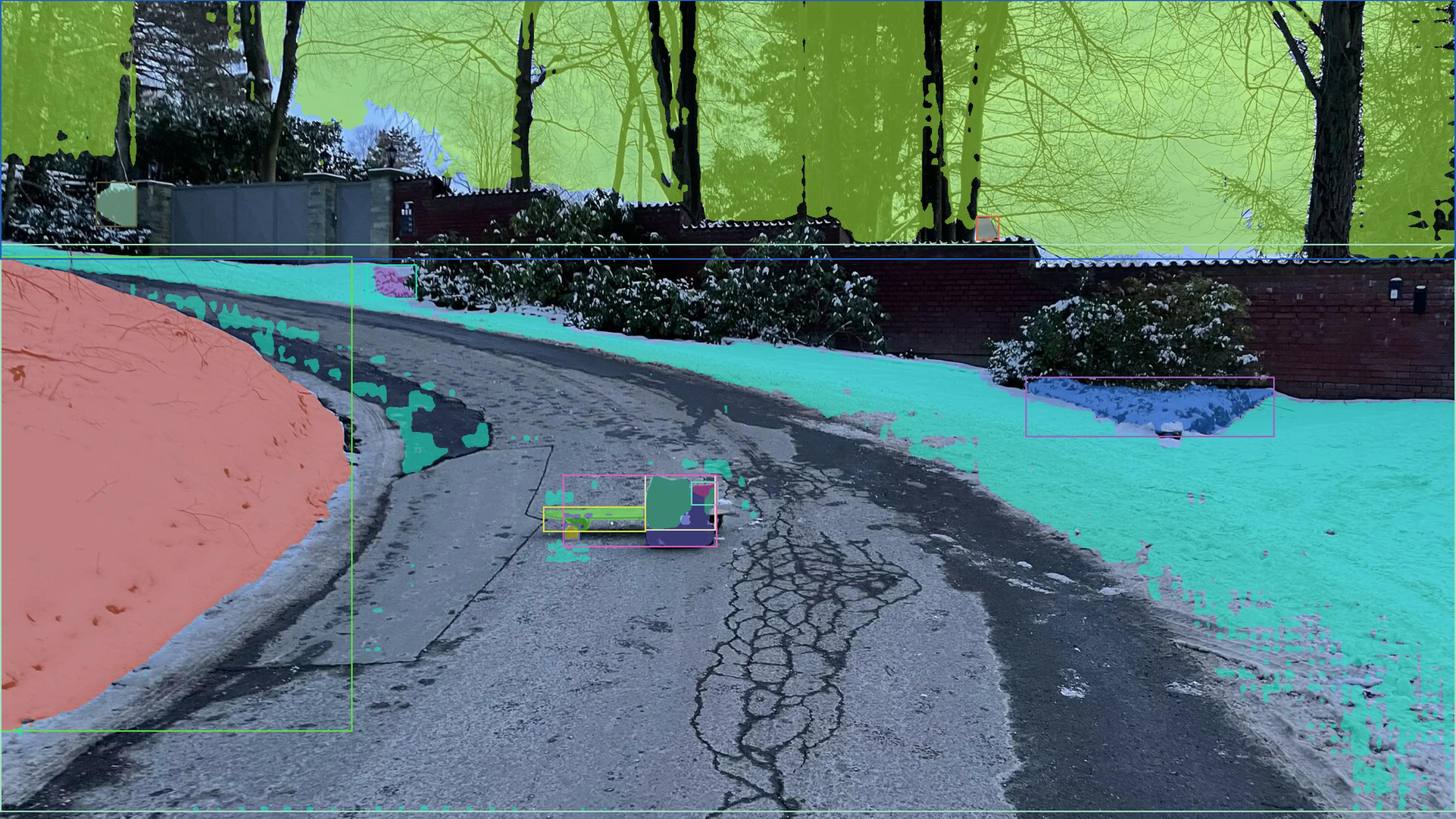}};
    \end{tikzpicture}}
    \caption{Qualitative comparison of all three methods on all three datasets.}
    \label{fig:qualitative}
\end{figure*}

%% file: tables/size_table.tex
\begin{table}[ht]
    \setlength{\tabcolsep}{4pt}
    \caption{
    Evaluation of three existing anomaly instance segmentation methods for three different buckets of object sizes in terms of AP and AP50. Higher scores indicate stronger performance.
    }
    \begin{tabularx}{\linewidth}{lYYYYYY}
    \toprule
    \multirow{2}{*}{Object Size} & \multicolumn{2}{c}{UGainS~\cite{nekrasov2023ugains}} & \multicolumn{2}{c}{M2A~\cite{rai2023mask2anomaly}} & \multicolumn{2}{c}{U3HS~\cite{gasperini2023u3hs}} \\
    \cmidrule(lr){2-3} \cmidrule(lr){4-5} \cmidrule(lr){6-7}
    & AP & AP50 & AP & AP50 & AP & AP50 \\
    \midrule
    $<1,\!000$ pix & 8.50 & 19.94 & 1.32 & 2.68 & 0.15 & 0.43 \\
    $1,\!000$-$10,\!000$ pix & 34.43 & 55.32 & 30.29 & 53.24 & 0.08 & 0.31 \\
    $>10,\!000$ pix & 17.41 & 31.69 & 32.19 & 48.25 & 0.05 & 0.15 \\
    \midrule
    Mean & 25.19 & 42.81 & 13.73 & 24.30 & 0.19 & 0.58 \\
    \bottomrule
    \end{tabularx}
    \label{tab:size}
\end{table}

%% file: sections/5_conclusion.tex
\section{Conclusion}
\label{sec:conclusion}
Detecting and accurately segmenting anomaly instances on roads is a significant challenge, requiring an understanding of 'objectness' without direct training on specific anomaly classes.
In this work, we introduced new benchmarks for anomaly instance segmentation and anomalous object detection that integrate three popular anomaly datasets with new instance-wise annotations. This is complemented with an evaluation protocol that treats large and small objects as equally important.
Our unified benchmark, termed OoDIS, provides a diverse set of anomalies that vary in size, rarity and the visual conditions in which they are presented. Comprising three datasets, OoDIS constitutes a challenging setting with a substantial number of images and annotation detail.
We evaluate the performance of current methods for segmenting anomaly instances and provide an intuition behind the results.
Our results show that current techniques struggle particularly with distant and small objects, and with precise segmentation masks.
The benchmark results suggest strong opportunities for advancement in the area.
As autonomous vehicle technologies continue to evolve, driven by large amounts of data, it remains a challenge to capture all possible real-world situations.
Our work addresses the need to evaluate instance segmentation and object detection as a step towards reliable autonomous driving and robot navigation.